\documentclass[journal]{IEEEtran}
\usepackage{times}
\usepackage{epsfig}
\usepackage{graphicx}
\usepackage{amssymb}
\usepackage{mathrsfs}
\usepackage{multirow}
\usepackage{subfigure}
\usepackage{array}
\usepackage{float}
\usepackage[linesnumbered,ruled,vlined]{algorithm2e}
\usepackage{algpseudocode}
\usepackage{amsmath}
\usepackage{booktabs}
\usepackage{dblfloatfix}
\usepackage{tabu}

\usepackage[pagebackref=true,breaklinks=true,colorlinks,bookmarks=false]{hyperref}

\ifCLASSINFOpdf
\else
\fi

\begin{document}
\title{SCIDA: Self-Correction Integrated Domain Adaptation from Single- to Multi-label Aerial Images}

\author{Tianze Yu$^\dagger$,~\IEEEmembership{Student Member,~IEEE,}
        Jianzhe Lin$^\dagger$,~\IEEEmembership{Student Member,~IEEE,}
        Lichao Mou,
        Yuansheng Hua,
        Xiaoxiang Zhu,~\IEEEmembership{Senior Member,~IEEE,}
        Z. Jane Wang,~\IEEEmembership{Fellow,~IEEE}
\thanks{$\dagger$ indicate equal contribution. Jianzhe Lin, Tianze Yu, and Z. Jane Wang are with the Department
of Electrical and Computer Engineering, University of British Columbia, Vancouver, BC, Canada. e-mail: jianzhelin, tianzey, zjanewang@ece.ubc.ca.}
\thanks{Lichao Mou, Yuansheng Hua, Xiaoxiang Zhu are with the Technical University of Munich and the German Aerospace Center, Germany.e-mail: lichao.mou,Yuansheng.Hua, Xiaoxiang.Zhu@dlr.de.}
}

\markboth{} 
{Shell \MakeLowercase{\textit{et al.}}: Bare Demo of IEEEtran.cls for IEEE Journals}

\maketitle

\begin{abstract}
Most publicly available datasets for image classification are with single labels, while images are inherently multi-labeled in our daily life. 
Such an annotation gap makes many pre-trained single-label classification models fail in practical scenarios. This annotation issue is more concerned for aerial images:  
Aerial data collected from sensors naturally cover a relatively large land area with multiple labels, while annotated aerial datasets, which are publicly available (e.g., UCM, AID), are single-labeled. As manually annotating multi-label aerial images would be time/labor-consuming, we propose a novel self-correction integrated domain adaptation (SCIDA) method for automatic multi-label learning.  SCIDA is weakly supervised, i.e., automatically learning the multi-label image classification model from using massive, publicly available single-label images. 
To achieve this goal, we propose a novel Label-Wise self-Correction (LWC) module to better explore underlying label correlations. 
This module also makes the unsupervised domain adaptation (UDA) from single- to multi-label data possible. 
For model training, the proposed model only uses single-label information yet requires no prior knowledge of multi-labeled data; and it predicts labels for multi-label aerial images. 
In our experiments, trained with single-labeled MAI-AID-s and MAI-UCM-s datasets, the proposed model is tested directly on our collected Multi-scene Aerial Image (MAI) dataset.
The code and data are available on GitHub(\url{https://github.com/Ryan315/Single2multi-DA}).
\end{abstract}


%
\IEEEpeerreviewmaketitle

\section{Introduction}
\begin{figure}[!htp]
\centering
\includegraphics[width=\linewidth]{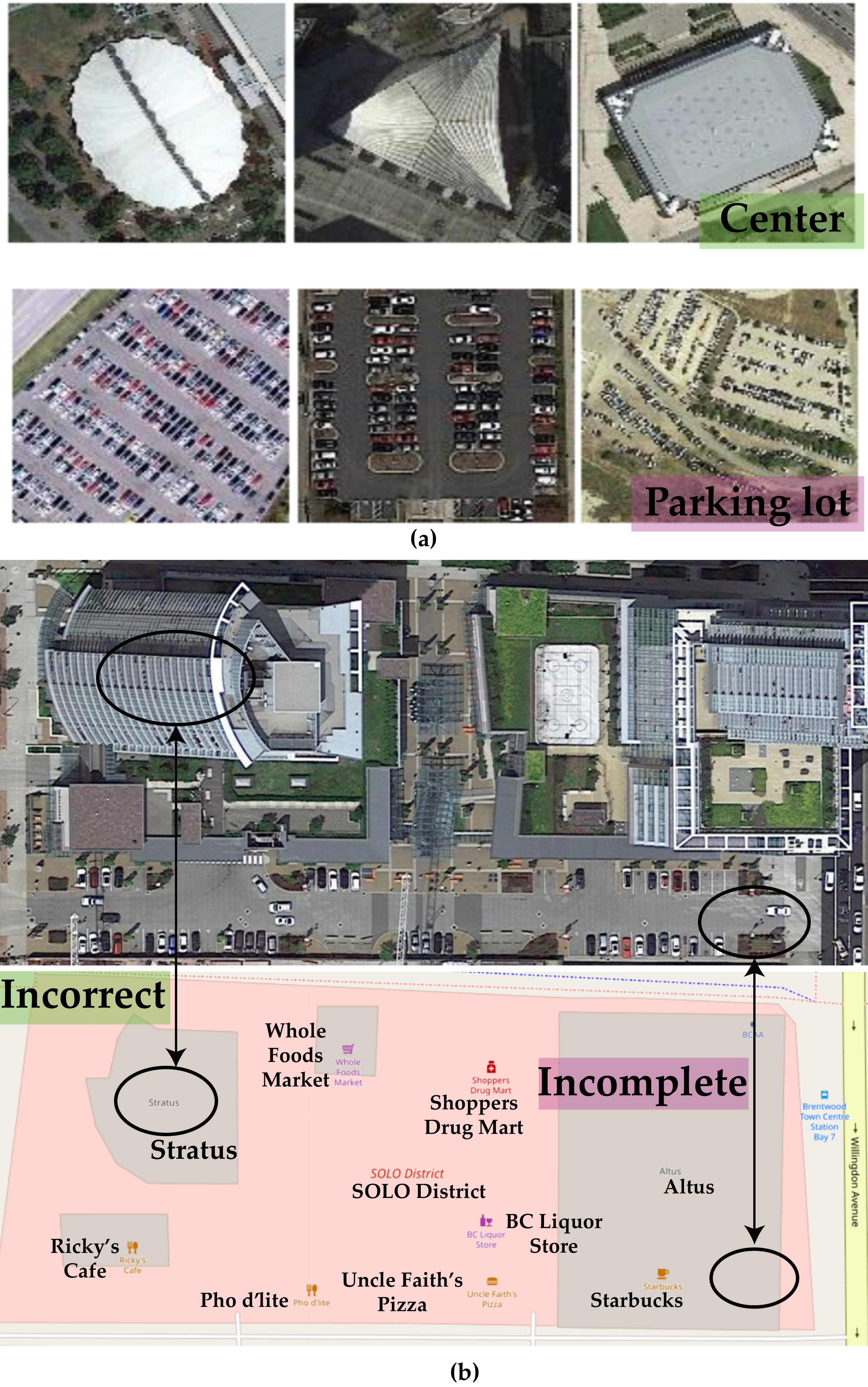}
\caption{(a) Single-label aerial image examples from the MAI-AID-s dataset, which serves as the source domain data. (b) The top is a multi-label aerial image from the MAI dataset, which serves as the target domain data. The bottom shows the corresponding noisy annotations from OSM for the top image. As indicated,  ``Center'' is incorrectly annotated as ``Stratus'', and ``Parking lot'' is missed.}
\label{fig:1}
\end{figure}

\IEEEPARstart{W}{ith} easy access to increasing aerial data from satellites/ Unmanned Aerial Vehicles (UAVs), annotating the newly collected aerial data is of great importance. 
However, obtaining clean multi-label annotations manually for aerial data has long been a challenging task. 
A recent trend for aerial data annotation is resorting to crowdsourcing data, such as OpenStreetMap (OSM). OSM, an editable map, is built/annotated by volunteers from scratch. 
However, the quality of such a manually annotated map may not be satisfying. 
Incompleteness and incorrectness are two primary concerns, as illustrated in the example in Fig. \ref{fig:1}.

Instead of resorting to crowdsourcing data, recent advances in machine learning (ML) make automatic annotation of aerial images possible. 
To train a reliable multi-label aerial image classification framework, we need to (1) design an efficient ML model architecture; and (2) learn with massive annotated data. 
However, collecting such multi-label aerial images with an exhaustive, consistent list of annotations requires significant time and effort. 
Furthermore, there is almost no publicly available multi-label aerial image dataset, and the online annotations from OSM are not always reliable. 
Training the model with noisy labels from OSM could lead to poor classification performance. 
Therefore, direct training of a multi-label aerial image classification model remains challenging.

An alternative way is to train the model with single-labeled aerial image data, since annotating a single-label aerial image is much easier than a multi-label one. 
Moreover, there are publicly available single-labeled datasets, e.g., AID and UCM datasets, which have covered almost all classes of interest for aerial images. 
Therefore, they could provide intense supervision for training a multi-label classification model. 
However, using the model trained by single-label data to predict the labels for a multi-label image is a challenging task. 
This task is illustrated in Fig. \ref{fig:1}, where single-labeled data (the source domain) is shown in Fig. \ref{fig:1}(a), and the corresponding multi-label data including the same labels (the target domain) is shown in Fig. \ref{fig:1}(b). 
To realize multi-label aerial image classification using single-labeled data, we formulate the problem as a domain adaptation task and propose a novel framework where prior information from single-label data can be adapted to multi-label aerial images.

\begin{figure*}[!htp]
  \centering
  \includegraphics[width=\linewidth]{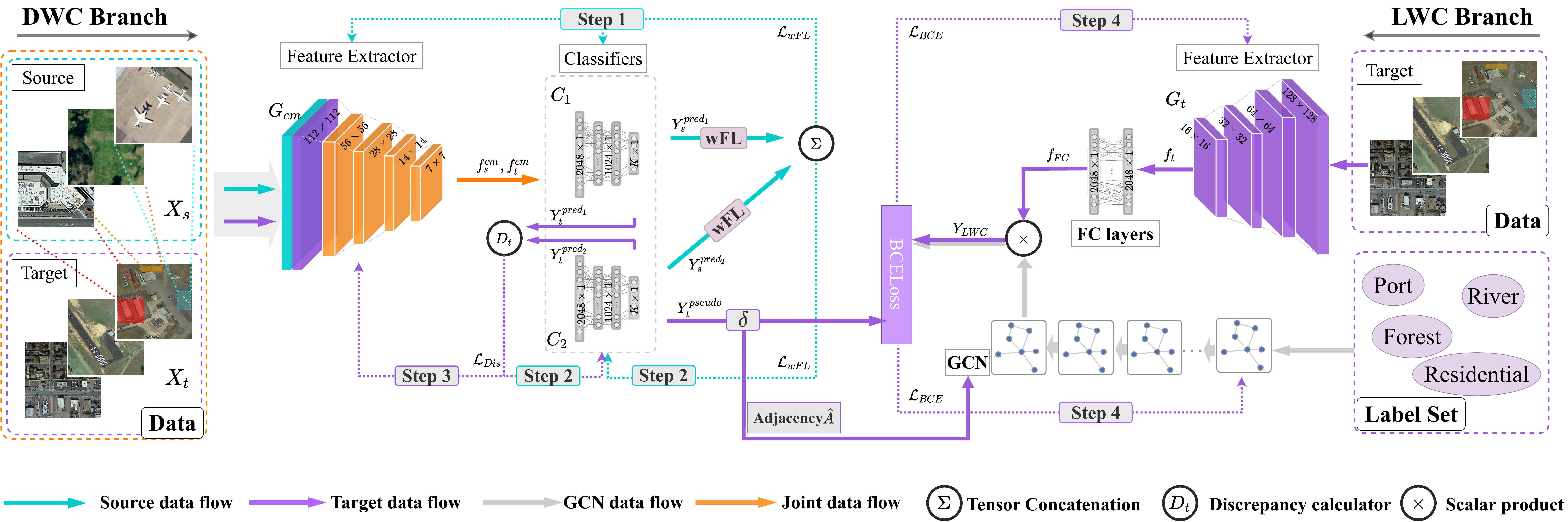}
  \caption{The flowchart of training the proposed SCIDA with the ResNet backbone. The flow mainly consists of the DWC branch and the LWC branch. $G_{cm}$ and $G_t$ are feature generators; and $C_1$ and $C_2$ represent classifiers. “wFL” means “weighted focal loss”, and $\delta$ controls the learning depth at each training step.}
\label{fig:mainflow}
\end{figure*}

To our knowledge, this is the first work to examine the challenging task of learning a multi-label aerial image classifier on large-scale datasets (the target domain) by using publicly available single-label data (the source domain). 
Our major contributions are as follows:
\begin{itemize}
    \item We propose a challenging single- to multi-label domain adaptation task. The target domain multi-label data are unannotated, large-scale, unconstrained aerial images in real-world scenarios, while the source domain is the annotated single-label aerial data publicly available (e.g., AID \cite{xia2017aid}, UCM \cite{yang2010bag}). 
    \item We propose a novel Self-Correction Integrated Domain Adaptation (SCIDA) framework, from single- to multi-label aerial images. Different from existing feature-based unsupervised domain adaptation methods, our framework explores the underlying label correlations by introducing the Label-Wise self-Correction (LWC) module. With GCN as the backbone, the LWC explores label correlations and iteratively corrects the pseudo labels learned by our domain adaptation module (the DWC branch, as in Fig. \ref{fig:mainflow}).
    \item We empirically compare labeling strategies for multi-label datasets to explore the learning potential of using single labels. Given the same annotation budget, our experiments show that the networks trained with single-label images can provide competitive performances as those learned with a fully annotated subset of multi-label images.       
    \item A new single-multi-label aerial image (MAI) dataset with clean labels is collected in our study to make the experiment possible. It will be the first web available dataset for multi-label aerial images. 
    
\end{itemize}

\section{Related Work}
\subsection{Partial Multi-Label Learning (PML)}
Multi-label learning has been an active research topic of practical importance, as images collected in the wild are always with more than one annotation \cite{tsoumakas2007multi}. 
Conventional multi-label learning research \cite{behpour2018arc}\cite{lyu2020partial}\cite{wang2018adaptive} mainly relies on the assumption that a small subset of images with full labels are available for training. 
However, this may be difficult to satisfy in practice, as manually yielded annotations always suffer from incomplete and incorrect annotation problems, as illustrated in Fig. \ref{fig:1}. 
Therefore, partial multi-label learning is emerging, which aims to learn a multi-label classification model from ambiguous data \cite{xie2020partial, sun2019partial, zhang2017disambiguation}. 
Traditional methods treat missing labels as negatives and wrong labels as positives during the model training process \cite{mahajan2018exploring, bucak2011multi, sun2017revisiting, wang2014binary, joulin2016learning}, while it could lead to degradation of the classification performance.

To mitigate the problem of missing or wrong labels, a novel approach for partial label learning is to treat missing labels as a hidden variable via probabilistic models and predict missing labels by posterior inference \cite{vasisht2014active, chu2018deep}. 
The work in \cite{misra2016seeing} models missing labels as negatives, and then corrects the induced error by learning a transformation on the output of the multi-label classifier. 
However, this approach requires high memory and is hard to optimize. 
Scaling these models to large datasets would be difficult \cite{deng2014scalable, huynh2020interactive}. 
Another recent trend for partial label learning can be found in \cite{dong2018learning, liu2006semi, feng2018leveraging}, which introduces curriculum learning and bootstrapping to increase the number of annotations. 
During model training, this approach uses the partially annotated data and the unannotated data whose predicted labels are with the highest confidences \cite{wang2019discriminative, zhang2020partial}. 
Curriculum learning is further combined with the graph model in \cite{durand2019learning} to better capture the label association when exploiting the unlabeled data during training. 
This approach still relies heavily on the unlabeled images, which would taint the training data when they are attached to incorrect labels. 
This problem is called semantic drift.

Different from the existing models, we propose a novel approach for multi-label learning with the extreme case of partial multi-label data, namely the \textbf{single label} data. 
Compared with commonly used partial multi-label data, single-label data are collected specifically for a single class. 
Therefore, such data are much easier to acquire online, and do not have the incomplete or incorrect annotation problem. 
In this paper, for the first time, we introduce domain adaptation to train the multi-label classification model using the annotated single-label data.

\subsection{Domain Adaptation (DA)}
Domain adaptation is a method to share knowledge between data from different datasets. 
DA aims to minimize the data gap between datasets \cite{ Revisiting2016Li, Autodial2017F, li2018adaptive}. 
Here, we formulate the knowledge transfer between the single-label data and multi-label data as a domain adaptation problem in our task. 
Recent years have witnessed the exploitation of adversarial domain adaptation, which stems from the technique proposed in \cite{dann}. 
The principal idea is to introduce adversarial learning by one feature generator and one domain discriminator \cite{dsn,coral, adda}. 
The generated features from the source domain and the target domain are aligned to confuse the domain discriminator until it cannot figure out which domain the features are from \cite{cycleda, gta, tamaazousti2019learning, zhu2019aligning}. 
One major limitation of existing adversarial domain adaptation methods is that they are not task-specific. 
For instance, the generated features in the DA model might not work well for the classifier \cite{Multi2018Long, Partial2018Long, CADA2018Long}. 
This problem could be even more severe for our task, since the classification tasks for the source and target domains are two different types (one is single-label classification, and the other is multi-label classification). 
A recent advance for task-specific DA is the Maximum Classifier Discrepancy (MCD) method, which is proposed to make the adversarial mechanism task-specific by constructing adversarial learning between \emph{task-specific classifiers} and \emph{feature generator} \cite{saito, lee2019sliced, kuroki2019unsupervised}. 
However, this method couldn't generalize well when the classifiers from two domains are not the same. 
To solve this problem, here we propose our new domain adaptation framework.

\section{Method}
In this section, we present the proposed SCIDA model for single- to multi-label domain adaptation.

\subsection{Overview}
In the proposed SCIDA framework, we need to correlate the domain-wise data, and also explore the label-wise correlation among target domain data. 
This label-wise correlation is used for self-correction in the framework. 

For the domain-wise correlation (DWC), we propose using the domain adaptation to correlate the two domains, due to the large domain gap between single-label and multi-label data. 
For the label-Wise self-Correction (LWC), due to the lack of correlation for the one-hot encoded source domain data, the LWC is learned with self-supervision in the target domain.
The Graph convolutional network (GCN) is introduced to model the LWC directly. 
The LWC is used for self-correction for multi-label classification.

The general flowchart of the proposed model is illustrated in Fig. \ref{fig:mainflow}, which is mainly made up of the DWC branch and the LWC branch. 
The inputs of the proposed framework are introduced as follows. 
The annotated source domain data is represented with $X_s = \{x_s^i, \textbf{y}_s^i\}_{i=1}^{N_s}$ (x and y represent the data and the label respectively), while the target domain data is represented with $X_t = \{x_t^i\}_{i=1}^{N_t}$ where $N$ represents the number of images in the dataset.


\subsection{Domain-wise Correlation}
For the domain-wise correlation, the goal of this branch is to align the features from source and target domains by utilizing two task-specific classifiers as a discriminator. 
The output of this branch is the pseudo label for target domain data. 
This branch is made up of three parts: a common feature extractor $G_{cm}$, two classifiers $C_1$ and $C_2$, and a classifier discriminator $D_t$. 
The extracted source domain feature $ f_s^{cm} = {G_{cm}}(X_s; {\theta}_{G_{cm}})$ and target domain feature $ f_t^{cm} = {G_{cm}}(X_t; {\theta}_{G_{cm}})$ are the inputs to the two classifiers. 
We need to detect target samples that largely deviate from the distribution of the source data, and align features from the two domains. 
As two classifiers are assumed to be effective on source domain samples with full annotations, the classification results from these two classifiers should be the same. 
While the target samples deviate from the source data distribution, and are likely to be classified differently by the two distinct classifiers. 
Our goal is to minimize the performance gap between the two classifiers for target domain samples. 
In our framework, this discrepancy for the two classifiers can be calculated in the target domain by $D_{t}$. 
If the discrepancy is minimized, we assume the data feature from the target domain is aligned with the source domain. 
And then we can use the source domain annotations to supervise the classification of the target domain data. 
The general training of this branch can be found in Sec. \ref{sec:ModelTraining}.

However, different from regular task-specific domain adaptation in \cite{saito}, the major challenge for domain adaptation from single-label to multi-label is the sharing of classifiers. 
If the single-label classifier is trained to generate multiple labels by setting a threshold for probability, the single-label source data will suffer from the problem of imbalanced training data. 
The imbalance issue contains two aspects: the imbalance of positive and negative samples in each class, and the imbalance of the samples in different classes in the entire dataset.
In this task, there are much fewer positive samples than the negative samples, with a ratio of about $1/K$ ($K$ is the number of classes).
To overcome this problem, we propose to use the \textbf{weighted focal loss (wFL)} instead of the regular cross-entropy loss for the optimization of the model, which is formulated as below:
\begin{equation}
  \begin{aligned}
{\cal L} =  & - \sum\limits_{i = 1}^K {{p_\beta }} (\alpha {y^i}{(1 - {p^i}({y^i}\left| {{x}} \right.))^\gamma }\log {p^i}({y^i}\left| {{x}} \right.) \\
    &+ (1 - \alpha )(1 - {y^i})p_i^\gamma ({y^i}\left| {{x}} \right.)\log (1 - {p^i}({y^i}\left| {{x}} \right.))), \label{eq:0}
  \end{aligned}
\end{equation}
where $p_\beta$ is the proportion of class-wise samples with respect to all the data in the dataset and fulfills $p_\beta \in (0,1)$ \& $\sum_{\beta=1}^C p_{\beta}=1$. $\alpha$ and $\gamma$ here are empirically set as 0.25 and 2 respectively \cite{lin2017focal}.

For the multi-label target dataset, the imbalance is eased to some extent. 
However, as annotations are not available for the target domain data, the classification in the target domain doesn't generate a loss for the supervision of the model. 
Instead, the discrepancy between the two classifiers in the target domain is regarded as the loss and used to optimize the model. 

\begin{figure}[b]
  \centering
  \includegraphics[width=\linewidth]{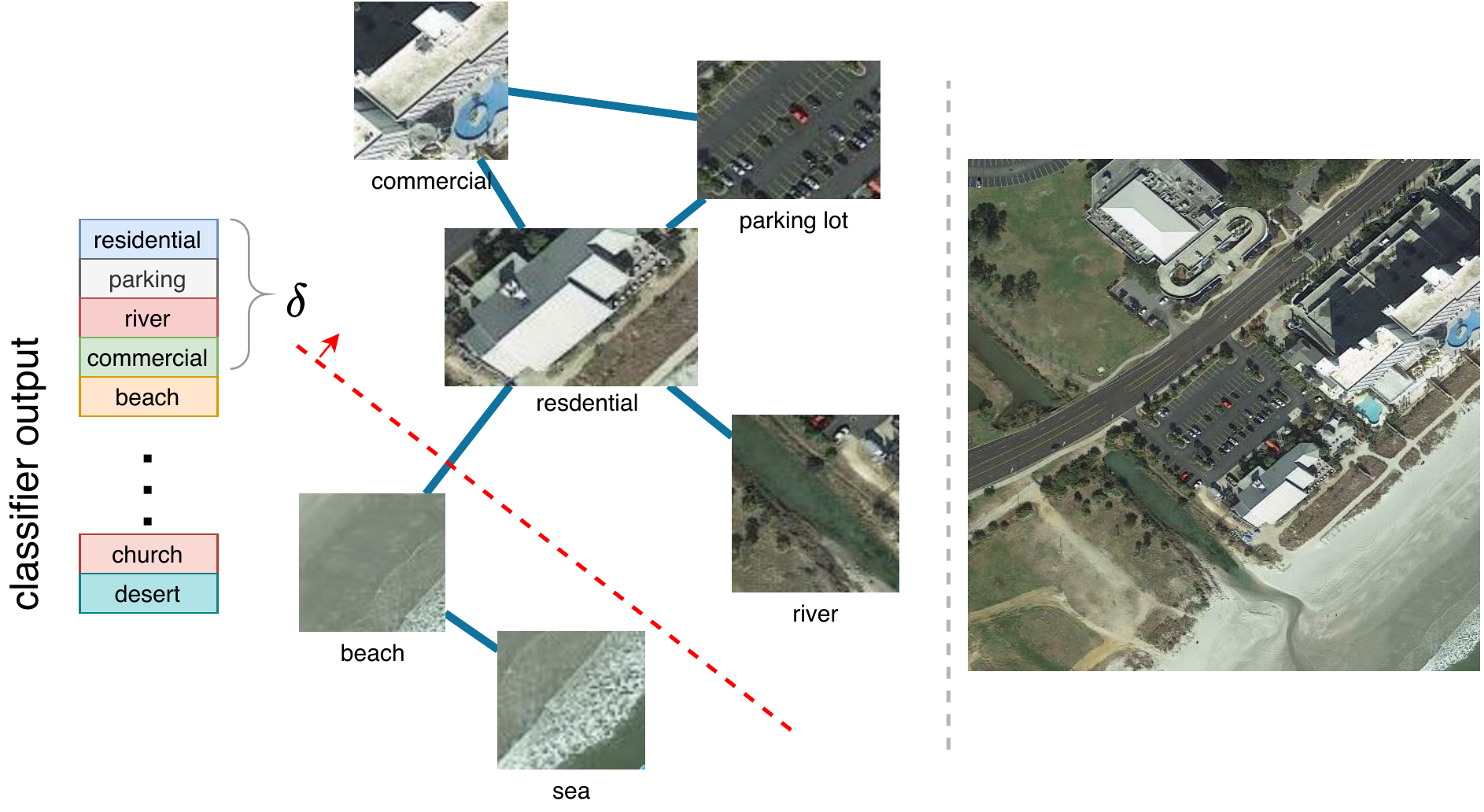}
  \caption{An example of $\delta$. The right shows the image, and the left shows the predicted labels.}
\label{fig:correlation}
\end{figure}

\subsection{Label-wise Correlation}
For the Label-Wise self-Correction (LWC) branch, the goal of this branch is to self-correct the pseudo label generated by the DWC branch. 
This branch is made up of two components, including a separate target convolution neural network with $G_t$ being the backbone for image classification, and a graph convolution network being the backbone for label correlation learning. 

\begin{algorithm}[t]
\SetAlgoLined
\scriptsize
\textbf{Stage I: Calculate the co-occurrence label} \\
 \textbf{Input:} Pseudo ground truth $Y_t^{pseudo}$\\
 \textbf{Output:} Correlation matrix A;\\
 \While{epoch $\leq$ max epoch}{
  \For{$batch\gets1$ \KwTo N}{
	\For{idx\_i, idx\_j $\gets1$ \KwTo num\_categories}{
	\If{ $Y_t^{pseudo}[idx\_i]$ \& $Y_t^{pseudo}[idx\_j]$ $\neq$ 0}{
		cor\_count[idx\_i][idx\_j] += 1
		}
   }
 }
 }
 \textbf{Stage II: Correlation matrix normalization} \\
\For{idx $\gets1$ \KwTo num\_categories}{
 	cor\_count[idx] = $\frac{cor\_count[idx]}{\sum{cor\_count[idx][:]}}$
 }
\Return cor\_count
\caption{Correlation matrix construction.}
\label{alg:cor_mat} 
\end{algorithm}

\subsubsection{Correlation matrix construction}
Label-wise correlation works by propagating label information between nodes based on the correlation matrix.
It's a crucial point of how to construct the correlation matrix.
As in the proposed unsupervised scenario, there's no pre-defined correlation matrix in the target task.
We will construct the correlation matrix in a target-data-driven approach based on the pseudo ground truth labels.
The procedure of constructing the correlation matrix is detailed in Alg. \ref{alg:cor_mat}.
For the pseudo ground truth, we introduce a hyper-parameter $\delta$ for controlling the number of pseudo ground truth labels for each image, as shown in Fig. \ref{fig:correlation}. 
As there's no prior label knowledge for the target domain data and the total number of labels for each image varies, we assume that the percentage of pseudo ground truth labels per image is a fixed constant $\delta$ (e.g., suppose the dataset has 20 labels in total, with $\delta =0.2$, the number of pseudo ground truth labels per image is 4). 
$\delta$ is used to determine how many nodes in GCN need to be updated in each iteration. 
An intuition is that $\delta$ is set as the average number of labels per image in the pseudo ground truth. 
Besides, as LWC branch is supervised by the pseudo labels generated by DWC branch, introducing $\delta$ will help to smooth the decision boundary of the classification.
In this way, the LWC could converge towards the correct direction, and meanwhile is not restricted to the output of DWC branch.
What's more, when calculating the co-occurrence of labels, some rare co-occurrence in the labels will introduce much noise and cause a long-tail distribution problem.
Introducing the parameter $\delta$ could solve the long-tail distribution problem to a great extent.
In the ablation study, we also analyzed the effects of the parameter $\delta$. 

\subsubsection{Label-wise GCN}
The inputs of this branch include four parts: the original target domain data, the target domain label set, the pseudo ground truth, as well as the normalized adjacency matrix $\hat{A}$ learned from the DWC branch (which represents the occurrence frequency of each label). 
We also need to point out that the label in the GCN module is different in a conventional convolution neural network, and is intended to solve the problem under a non-Euclidean topological graph.
The computation graph is generated based on the label embedding of each node and its neighbors. 
We follow a common practice\cite{chen2019multi} to deploy GCN:
\begin{equation}
    H^{(l+1)} = \sigma(\hat{A}H^{(l)}W^{(l)})  
\end{equation}
where $\hat{A}$ is the normalized adjacency matrix mentioned above, $H^{(l)}$ denotes the label embedding at the $l$-th layer in GCN, $W^{(l)}$ is a learnable transformation matrix, and
$\sigma$ acts as a non-linear operation. We employ LeakyReLU to implement this operation.

The general operation routine for this branch is as follows: The feature vector of target domain data is first extracted by $ {G_t}(x_t; {\theta}_{G_t})$. 
This feature vector is fed to two fully connected (FC) layers and a $2048\times1$ feature vector $f_{FC}$ is generated. 
For the input of the GCN model, GloVe\cite{pennington2014glove} is used to generate the embedding of labels. 
Then the output of GCN (a $2048\times K$ matrix) and $f_{FC}$ are fed to a scalar product layer to produce a classification result $Y_{LWC}$ for this branch. 
The difference between the classification result $Y_t^{pred_2}$ from $C_2$ (we assume $C_1$ and $C_2$ can get a unified result finally) and $Y_{LWC}$ is used to generate a binary cross-entropy loss (BCE loss), which will optimize all components in the LWC branch.

\section{Two-stage Model Training}
\label{sec:ModelTraining}
In the training procedure, we learn the parameters of DWC and LWC branches jointly and iteratively. 
In the primary stage, we initialize the pseudo label by the DWC branch. In the second stage, both branches are trained to optimize the pseudo label. The iterative training way is introduced as follows, and concluded in Alg. \ref{alg:duan}.

\begin{algorithm}[t]
\SetAlgoLined
\scriptsize
 \textbf{Input:} $X_s, X_t, Y_s, \hat{A}$, labels; \\
 \textbf{Output:} Parameters for $G_{cm}$, $C_1$, $C_2$, GCN, and $G_t$;\\
 \While{epoch $\leq$ max epoch}{
  \For{$batch\gets1$ \KwTo N}{
   \textbf{Pseudo labels generation:} Input the normalized source and target domain data to optimize DWC branch by minimizing Eq. (\ref{eq:step1}), (\ref{eq:step2}), (\ref{eq:step3}), and generate $Y_t^{pseudo}$;
   \texttt{\\}
   \textbf{Self Correction:} Update both DWC and LWC by minimizing Eq. (\ref{eq:step4}), which indicates the difference between $Y_t^{pseudo}$ and $Y_{LWC}$.   
   }
 }
\caption{Training for SCIDA.}
\label{alg:duan} 
\end{algorithm}

\subsection{Pseudo labels generation in DWC} 
Pseudo labels generation is generated by an adversarial domain adaptation way in DWC branch \cite{saito}. 
We first initialize the weights of the feature generator $G_{cm}$ and the classifiers $C_1, C_2$, and freeze other components. The model in this stage is optimized by the weighted focal loss calculated on the annotated source domain data. This process can be formulated as:

\begin{equation}
  \min\limits_{\theta_{G_{cm}}, \theta_{C_1}, \theta_{C_2}} \mathcal{L}_{wFL}(X_s, \textbf{Y}_s)
  \label{eq:step1} 
\end{equation}
here $\mathcal{L}$ is defined in Eq. (\ref{eq:0}).

We initialize the parameters of $G_{cm}, C_1, C_2$ and train the framework for classification in the source domain, as well as to achieve an adversarial training ready state.
Then we use an adversarial manner to train the two classifiers $C_1$ and $C_2$ using the domain discriminator $D_t$, and optimize the weights of $G_{cm}$. 

To be more specific, we first maximize the discrepancy loss generated by $D_t$ for target domain data. 
This loss is used for optimizing $C_1$ and $C_2$. The purpose is to identify target samples whose extracted features deviate the most from the distribution of the source domain features. 
At the same time, we need to keep $C_1$ and $C_2$ effective for the classification of the source domain data. We formulate this by 

\begin{equation}
\mathop {\min }\limits_{{\theta _{{G_{cm}}}},{\theta _{{C_1}}},{\theta _{{C_2}}}} {{\cal L}_{wFL}}({X_s},{Y_s}) - {{\cal L}_{dis}}(Y_t^{pred_1},Y_t^{pred_2})
  \label{eq:step2}
\end{equation}
here $Y_t^{pred_1},Y_t^{pred_2}$ represent multi-label predictions from $C_1$ and $C_2$ respectively, and ${\cal L}_{dis}$ is computed with a scalar subtraction between these two \cite{saito}.

Then, adversarial to Eq. (\ref{eq:step2}), we minimize the classifier discrepancy loss by optimizing the weights of $G_{cm}$, in order to encourage uniformed classification results from the two classifiers. $C_1$ and $C_2$ are frozen now, and the objective function is 

\begin{equation}
  \min\limits_{\theta_{G_{cm}}} \mathcal{L}_{dis}(Y_t^{pred_1},Y_t^{pred_2})
  \label{eq:step3}
\end{equation}

Finally, we assume $Y_t^{pred}$ from $C_2$ to be the $Y_t^{pseudo}$. 
We need to point out that the $Y_t^{pred}$ from $C_1$ is equal to $Y_t^{pred}$ from $C_2$ ($Y_t^{pseudo}$), when the training converges. 

\subsection{Self Correction}
Self Correction is achieved unsupervised in the target domain for self-correcting the pseudo label $Y_t^{pseudo}$. Only by learning the label correlation with GCN (which can be understood as a self-supervision) but without extra supervisions/annotations, the $Y_t^{pseudo}$ is optimized and corrected.

In this stage, we first get $Y_{LWC}$ from the LWC branch for the target domain data. 
This obtained $Y_{LWC}$ from the scalar product operation (as in Fig. \ref{fig:mainflow}) is used for self-correction for the pseudo label $Y_t^{pseudo}$. 
Noted that here $Y_{LWC}$ is the label predictions of target domain samples from LWC branch. $Y_{LWC}$ can be represented as $Y_{LWC} = \{\hat{y}^1, \hat{y}^2,...,\hat{y}^N\}$, supposing there are N target domain samples. 
As we note that the choice of different losses in this stage doesn't make noticeable difference, we empirically choose the BCE loss as below:

\begin{equation}
  \min\limits_{\theta_{G_t}, \theta_{GCN}, \theta_{G_{cm}}, \theta_{C_1}, \theta_{C_2}} \mathcal{L}_{BCE}(Y_{LWC}, Y_t^{pseudo})
  \label{eq:step4}
\end{equation}
and $\mathcal{L}_{BCE}$ for sample $x_i$ is further defined as
\begin{equation}
  \begin{aligned}
{\mathcal{L}_{BCE}} = - \frac{1}{K}\sum\limits_{i = 1}^K( {y^i}\log (\hat{y}^i) + (1 - {y^i})\log (1 - (\hat{y}^i)))
    \end{aligned}
\end{equation}
$y^i$ is the pseudo labels from DWC branch. We need to point out that both $y^i$ and $\hat{y}^i$ are variables being optimized, and the final convergence is only realized when the $y^i$ equals to $\hat{y}^i$. 
This BCE loss will be used to optimize the parameters of the whole network, including both DWC and LWC branches. 


\begin{figure*}[!tbp]
\centering
\subfigure[]{
\begin{minipage}[t]{0.2\linewidth}
\centering
\includegraphics[width=\textwidth]{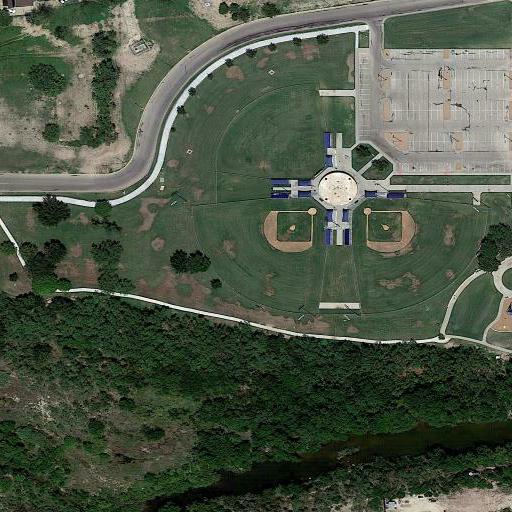}
\end{minipage}%
}%
\subfigure[]{
\begin{minipage}[t]{0.2\linewidth}
\centering
\includegraphics[width=\textwidth]{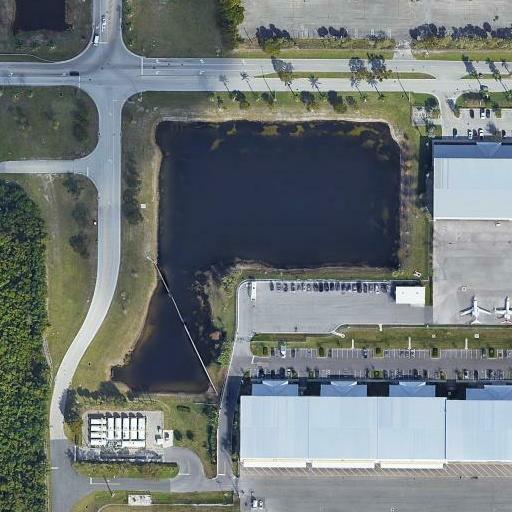}
\end{minipage}%
}%
\subfigure[]{
\begin{minipage}[t]{0.2\linewidth}
\centering
\includegraphics[width=\textwidth]{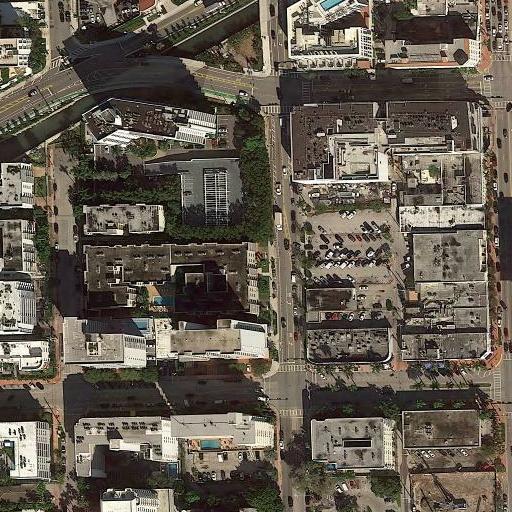}
\end{minipage}%
}%
\subfigure[]{
\begin{minipage}[t]{0.2\linewidth}
\centering
\includegraphics[width=\textwidth]{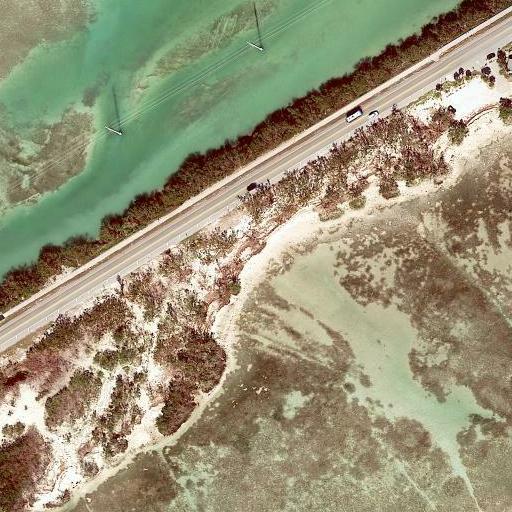}
\end{minipage}%
}%
\subfigure[]{
\begin{minipage}[t]{0.2\linewidth}
\centering
\includegraphics[width=\textwidth]{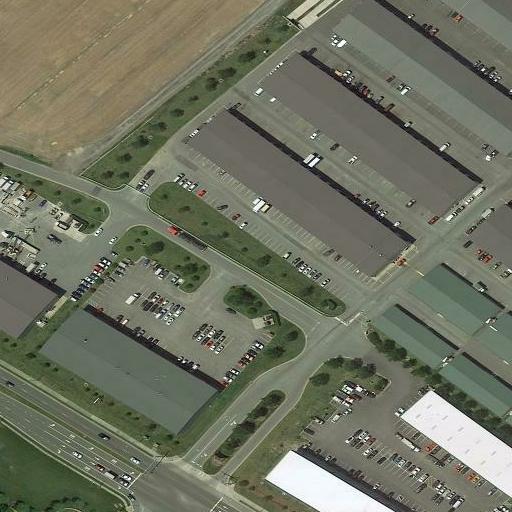}
\end{minipage}%
}%

\subfigure[]{
\begin{minipage}[t]{0.2\linewidth}
\centering
\includegraphics[width=\textwidth]{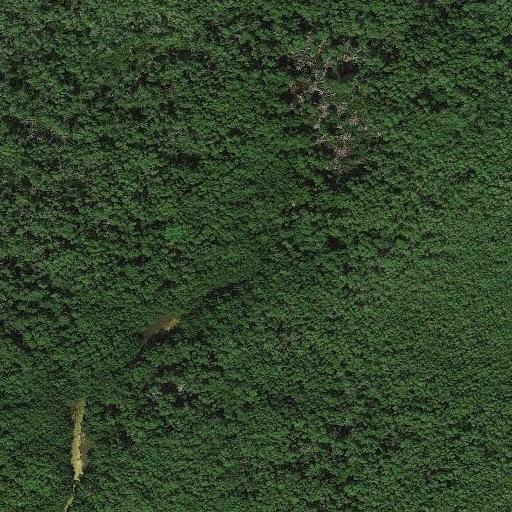}
\end{minipage}%
}%
\subfigure[]{
\begin{minipage}[t]{0.2\linewidth}
\centering
\includegraphics[width=\textwidth]{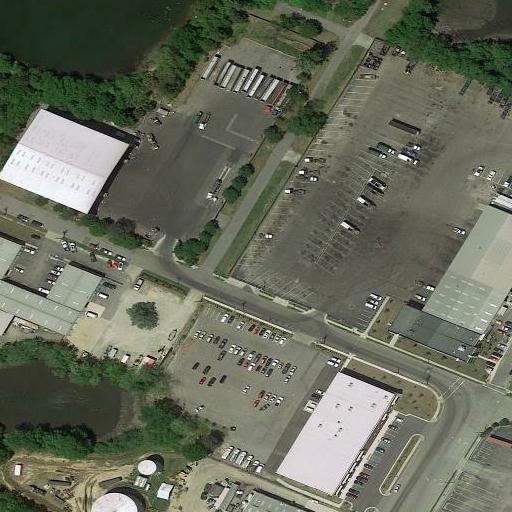}
\end{minipage}%
}%
\subfigure[]{
\begin{minipage}[t]{0.2\linewidth}
\centering
\includegraphics[width=\textwidth]{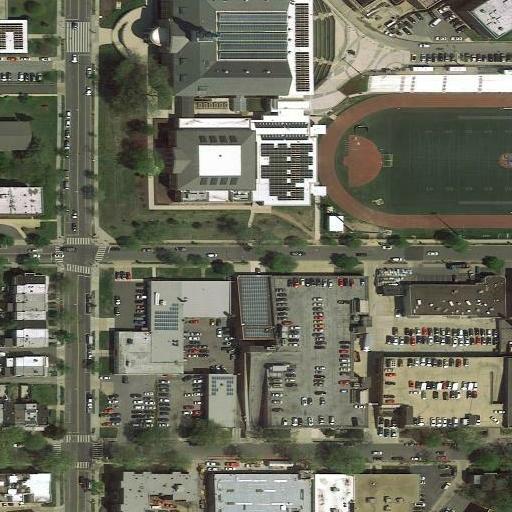}
\end{minipage}%
}
\subfigure[]{
\begin{minipage}[t]{0.2\linewidth}
\centering
\includegraphics[width=\textwidth]{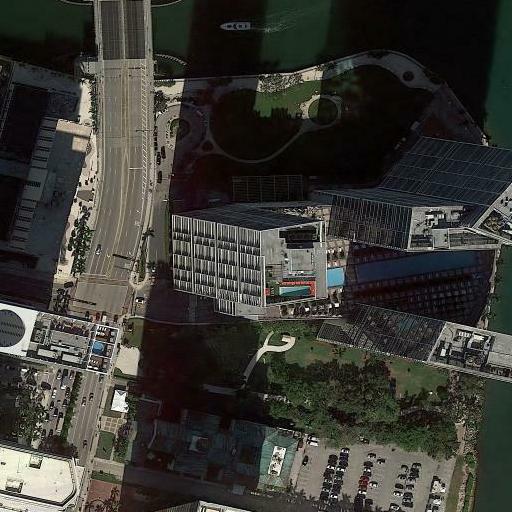}
\end{minipage}%
}%
\subfigure[]{
\begin{minipage}[t]{0.2\linewidth}
\centering
\includegraphics[width=\textwidth]{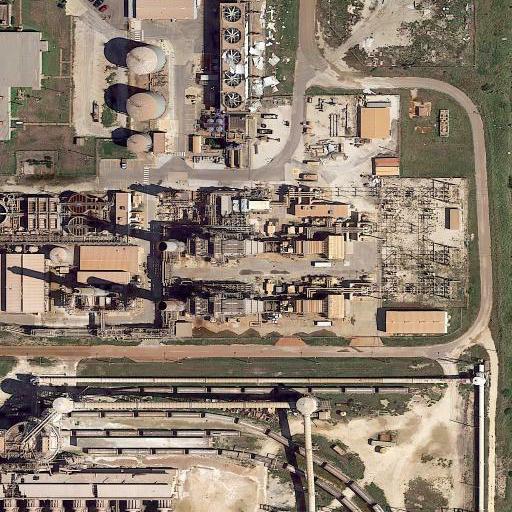}
\end{minipage}%
}%
\caption{Examples images with multiple labels from MAI-AID-m dataset. The labels are: (a) baseball, parking lot, park, river; (b) airport, forest, lake; (c) commercial, bridge, parking lot, residential; (d) beach, bridge; (e) farmland, commercial, parking lot; (f) forest, river; (g) commercial, lake, parking lot, residential, storage tank; (h) stadium, soccer, residential, parking lot; and (i) river, bridge, commercial.}
\label{fig:AID examples}
\end{figure*}

\begin{figure*}[!tbp]
\centering
\subfigure[]{
\begin{minipage}[t]{0.25\linewidth}
\centering
\includegraphics[width=\textwidth]{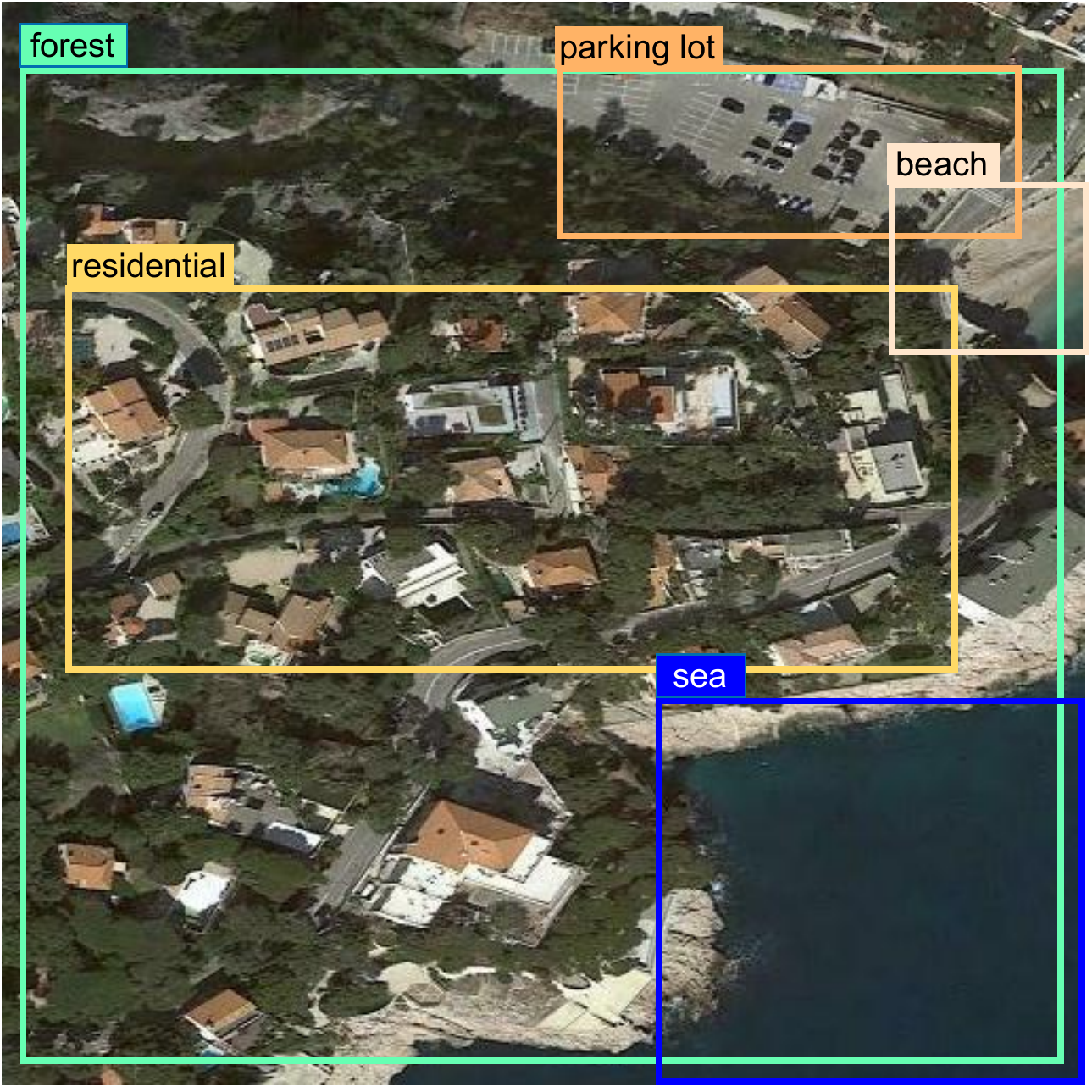}
\end{minipage}%
}%
\subfigure[]{
\begin{minipage}[t]{0.25\linewidth}
\centering
\includegraphics[width=\textwidth]{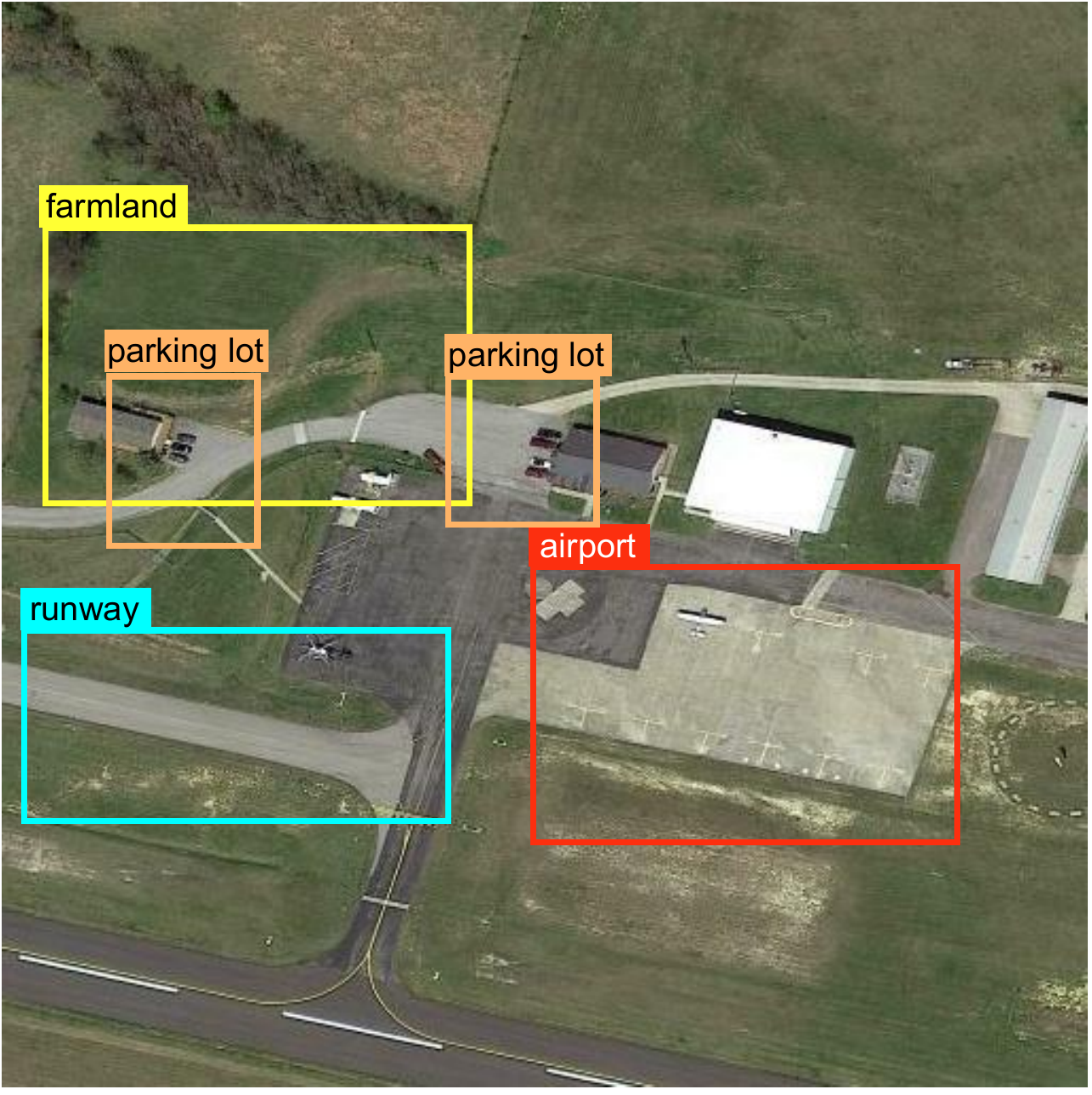}
\end{minipage}%
}%
\subfigure[]{
\begin{minipage}[t]{0.25\linewidth}
\centering
\includegraphics[width=\textwidth]{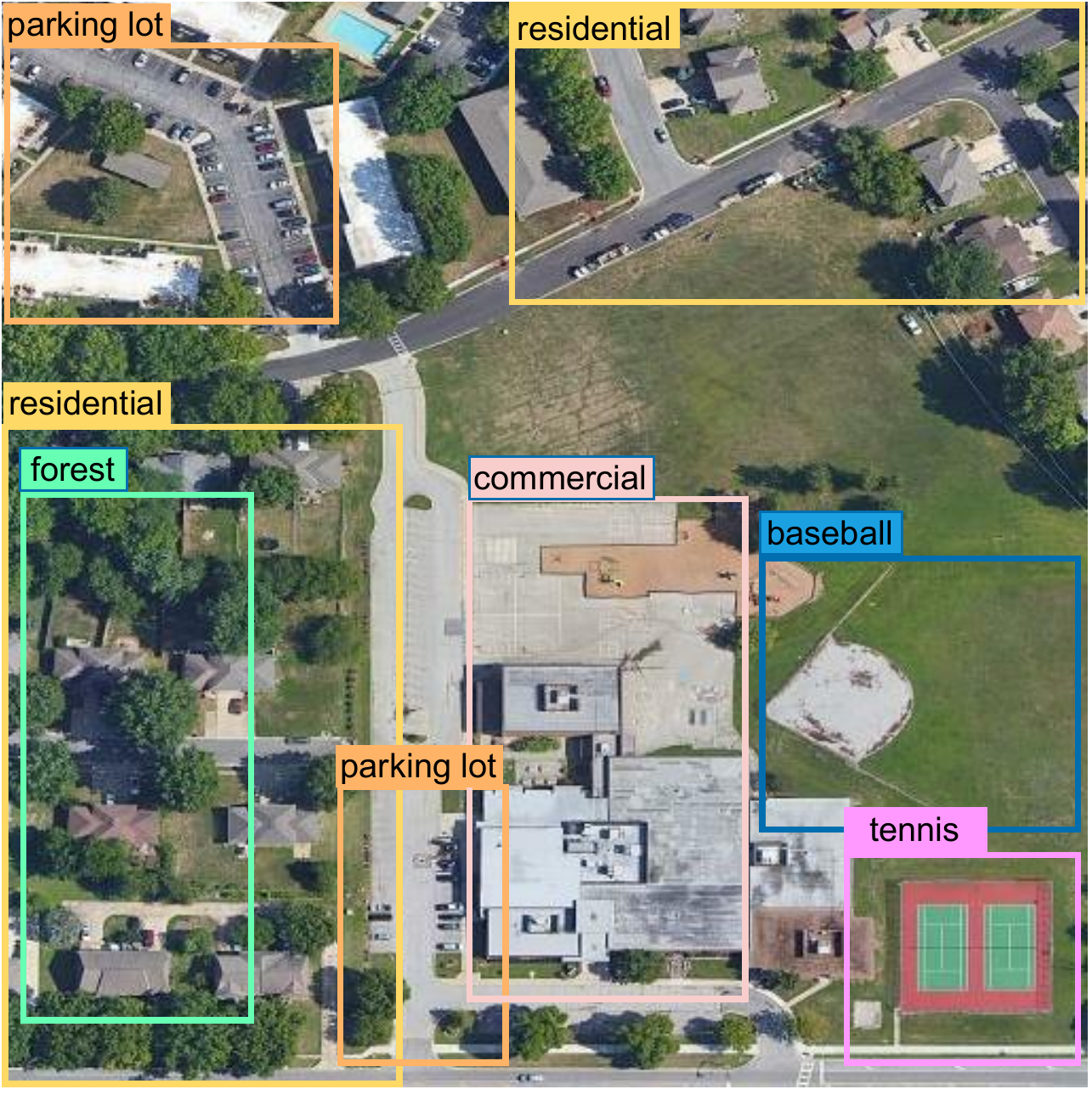}
\end{minipage}%
}%
\subfigure[]{
\begin{minipage}[t]{0.25\linewidth}
\centering
\includegraphics[width=\textwidth]{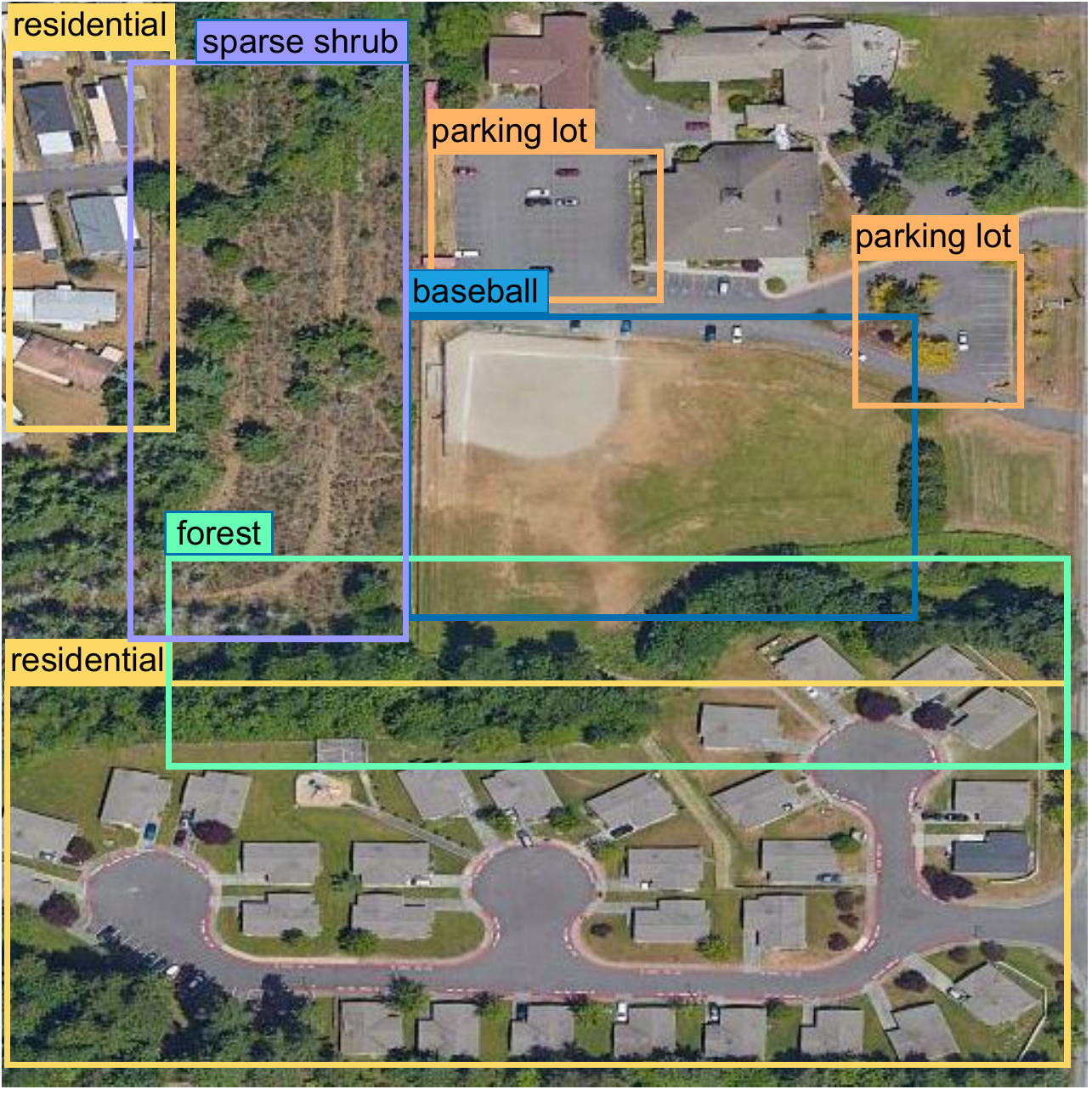}
\end{minipage}%
}%
\centering
\caption{Examples with object level annotations from the MAI-UCM-m dataset. The annotations are: (a) beach, sea, residential, forest, parking lot; (b) airport, farmland, parking lot, runway; (c) commercial, residential, forest, parking lot, baseball, tennis; and (d) sparse shrub, forest, baseball, residential, parking lot.}
\label{fig: UCM examples}
\end{figure*}

\begin{figure*}[!tbp]
\centering
\subfigure[Correlation visualization of MAI-AID-m.]{
\begin{minipage}[t]{0.5\linewidth}
\centering
\includegraphics[width=\textwidth]{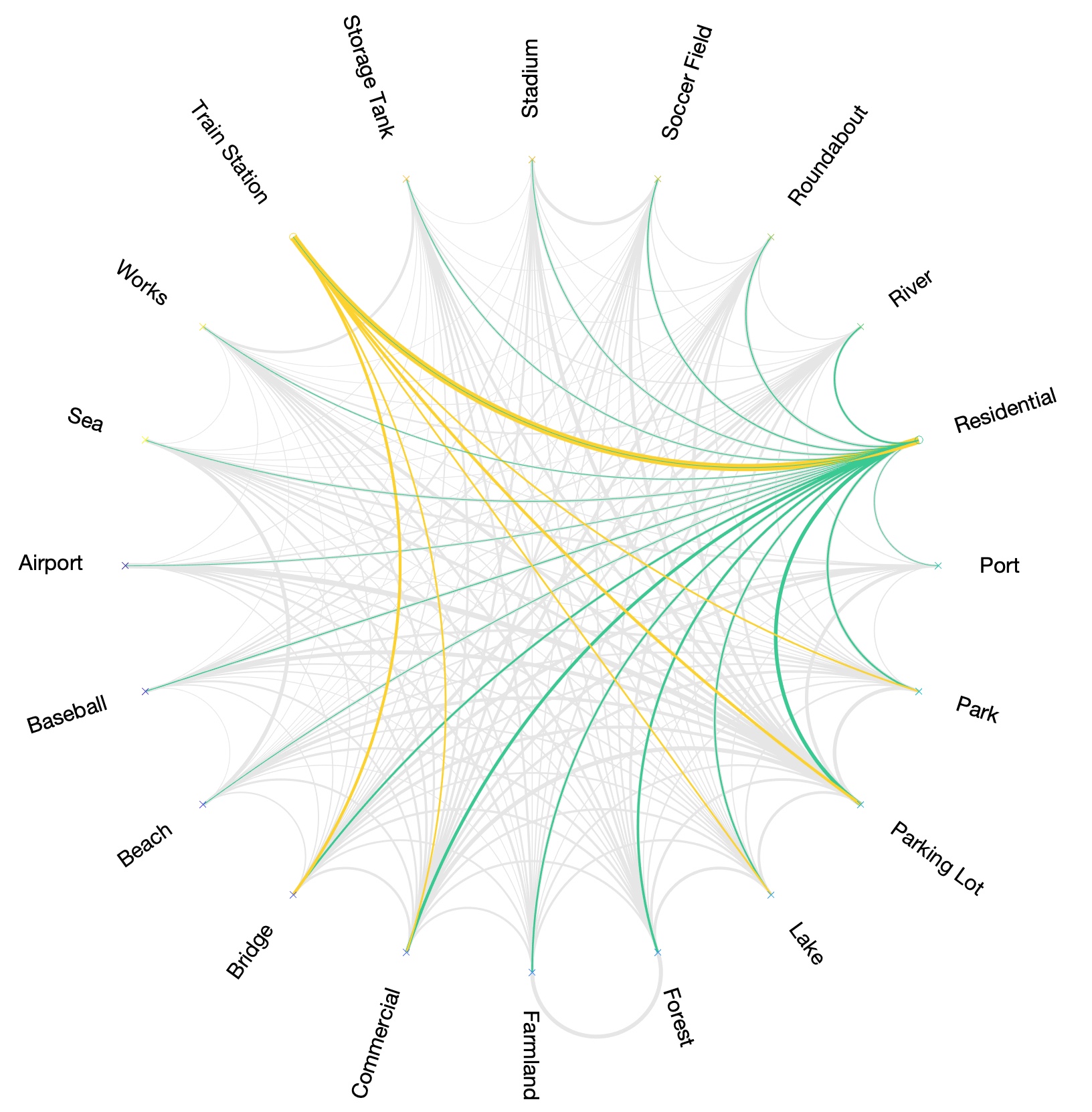}
\end{minipage}%
}%
\subfigure[Correlation visualization of MAI-UCM-m.]{
\begin{minipage}[t]{0.5\linewidth}
\centering
\includegraphics[width=\textwidth]{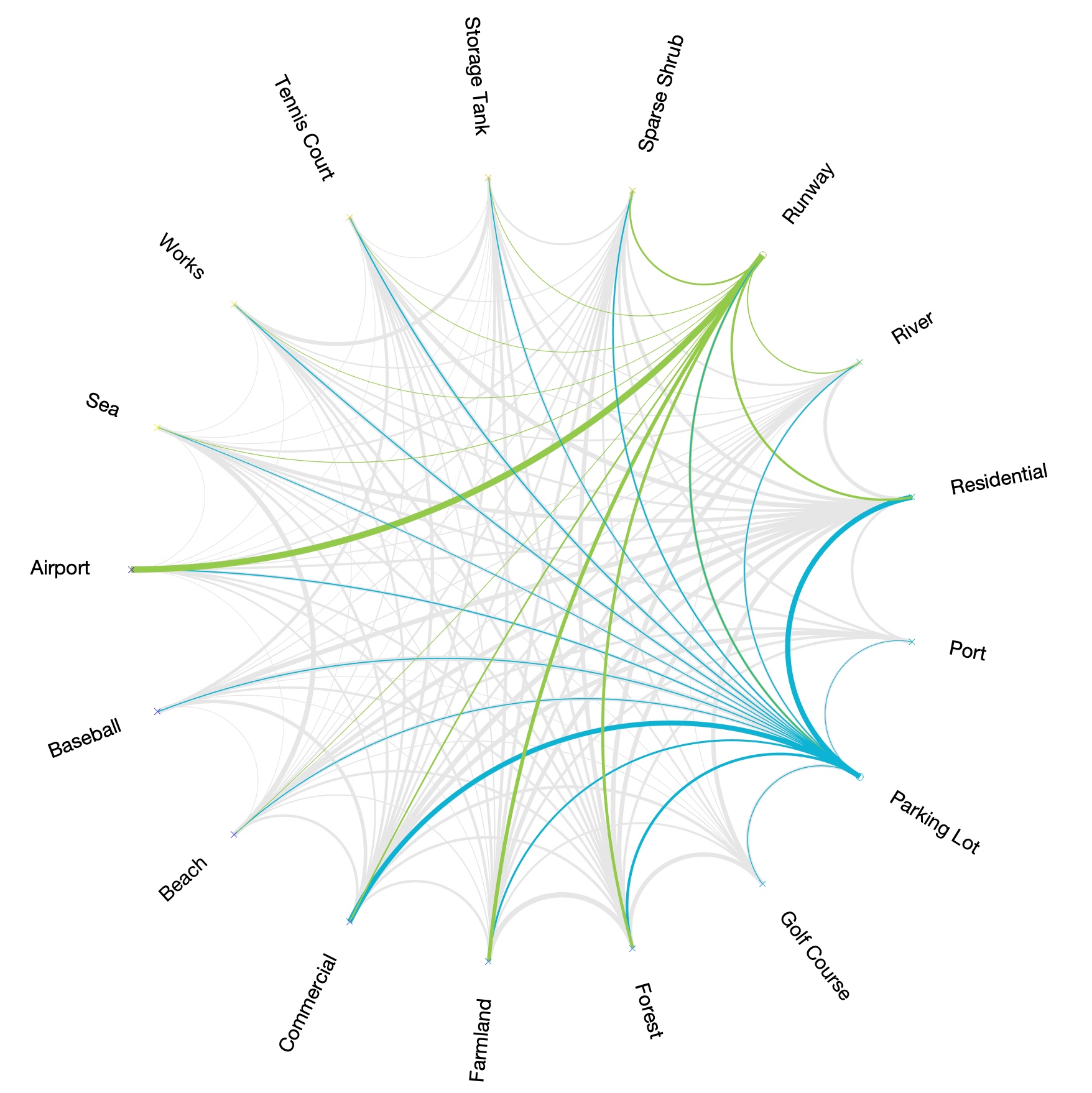}
\end{minipage}%
}%
\centering
\caption{Correlation visualization of the proposed MAI dataset. Here grey lines show all connections between label pairs; the linewidths represent the correlation values between label pairs; the categories with more and less occurrence are highlighted in different colors for demonstration. I.e., they are `Residential' and `Train Station' in MAI-AID-m, and they are `Parking Lot' and `Runway' in MAI-UCM-m respectively.}
\end{figure*}

These two stages will iterate until convergence in the training process. 
Finally, $Y_t^{pseudo}$ will stay unchanged, and we assume it as $Y_{pred}$ in the testing phase.

\section{Experiments}

\subsection{Datasets, Setup, and Evaluation Metrics}

\subsubsection{MAI Dataset}
\begin{figure*}[!tbp]
\centering
\subfigure[Annotation distribution of MAI-AID-m.]{
\begin{minipage}[t]{0.5\linewidth}
\centering
\includegraphics[width=\textwidth]{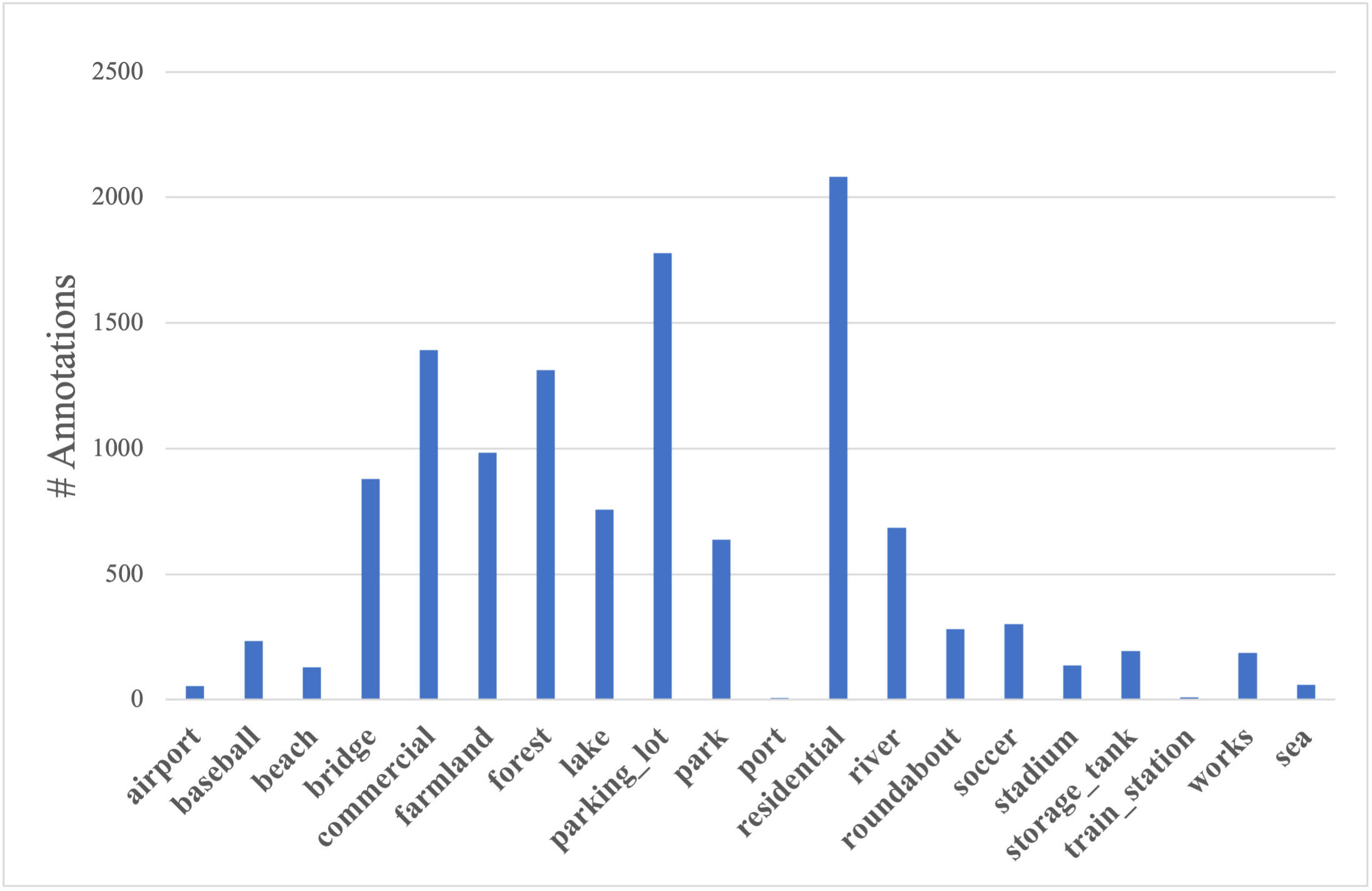}
\end{minipage}%
}%
\subfigure[Annotation distribution of MAI-UCM-m.]{
\begin{minipage}[t]{0.5\linewidth}
\centering
\includegraphics[width=\textwidth]{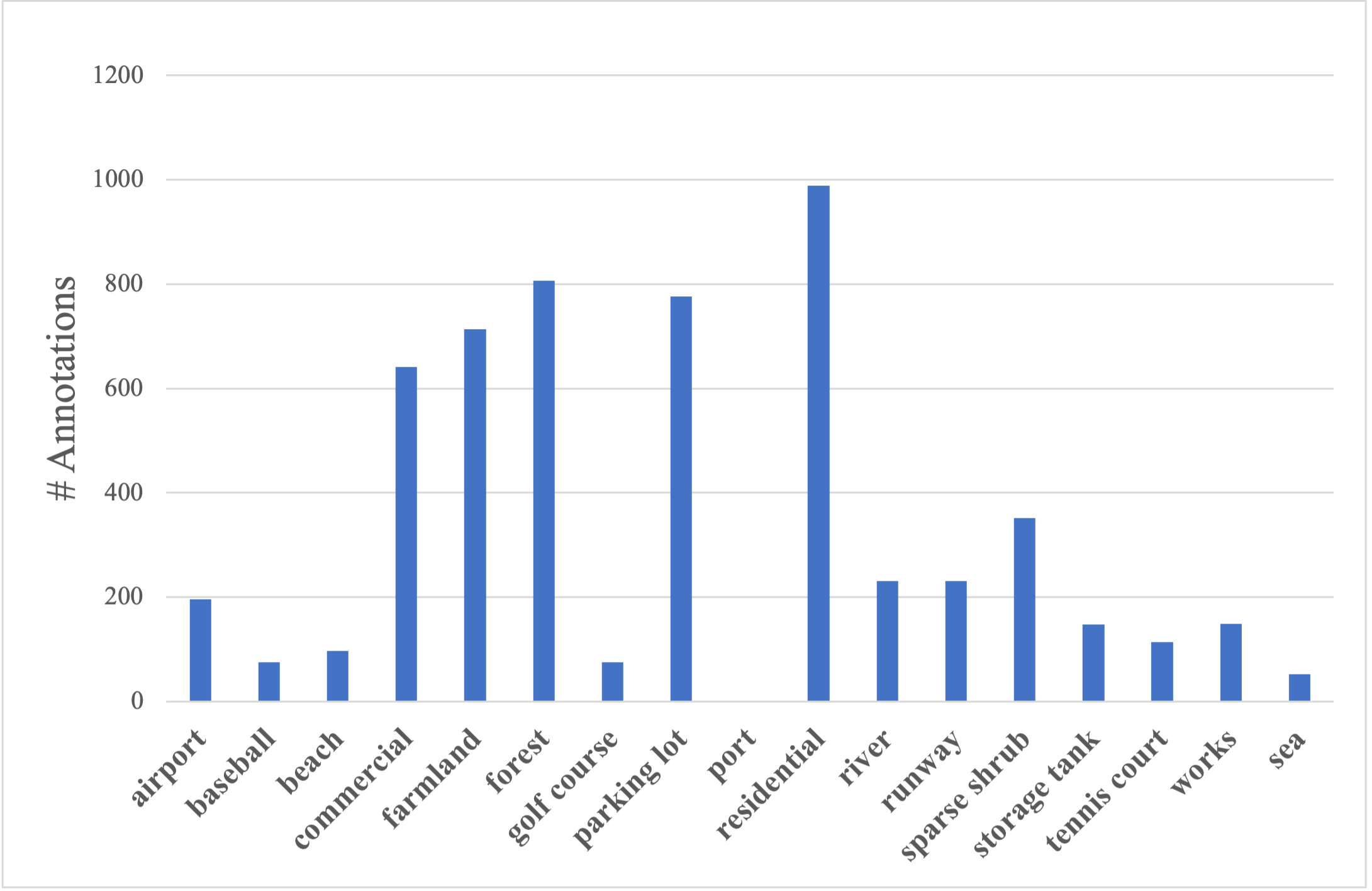}
\end{minipage}%
}%

\subfigure[Image distribution of MAI-AID-m.]{
\begin{minipage}[t]{0.5\linewidth}
\centering
\includegraphics[width=\textwidth]{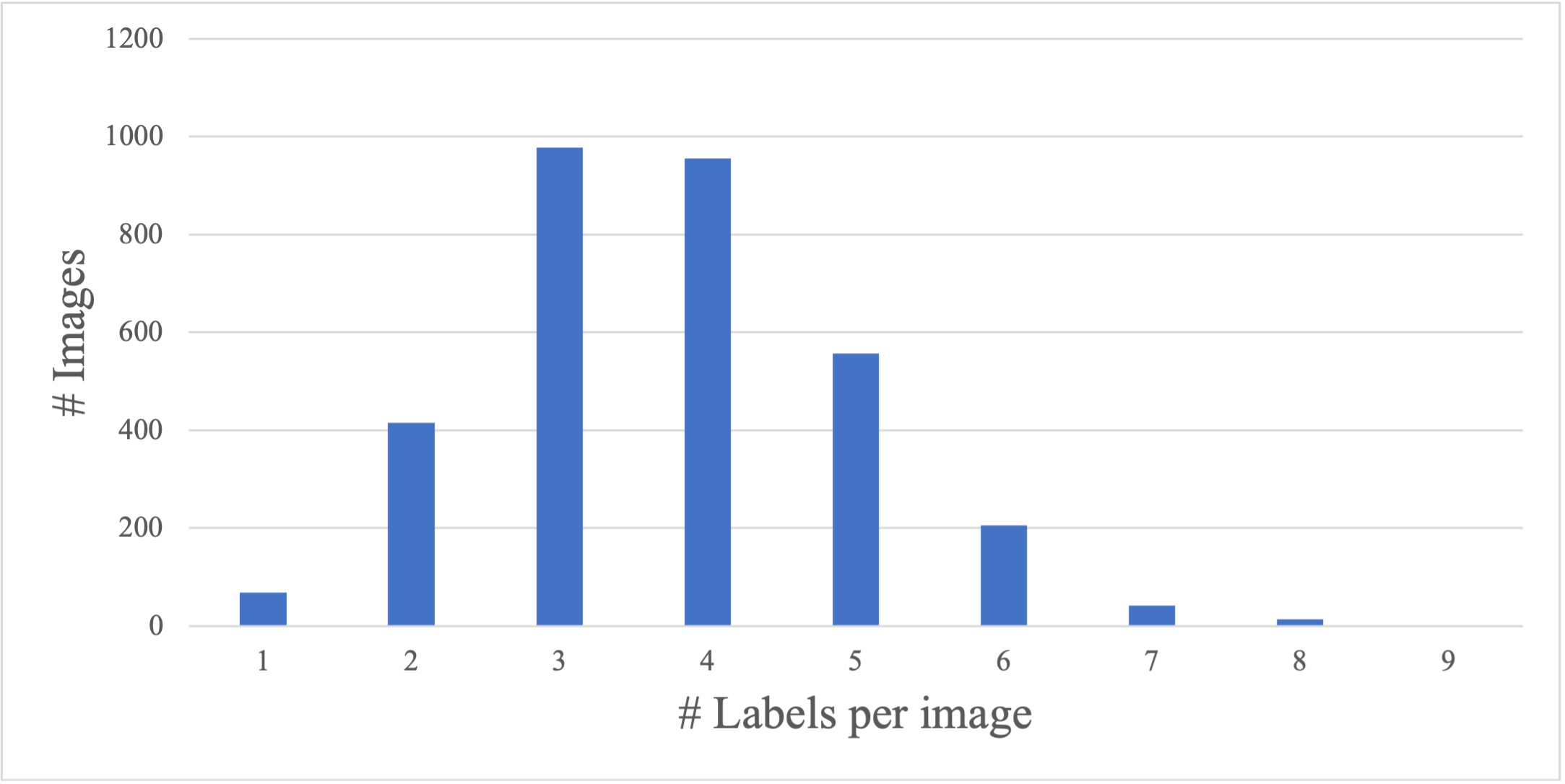}
\end{minipage}%
}%
\subfigure[Image distribution of MAI-UCM-m.]{
\begin{minipage}[t]{0.5\linewidth}
\centering
\includegraphics[width=\textwidth]{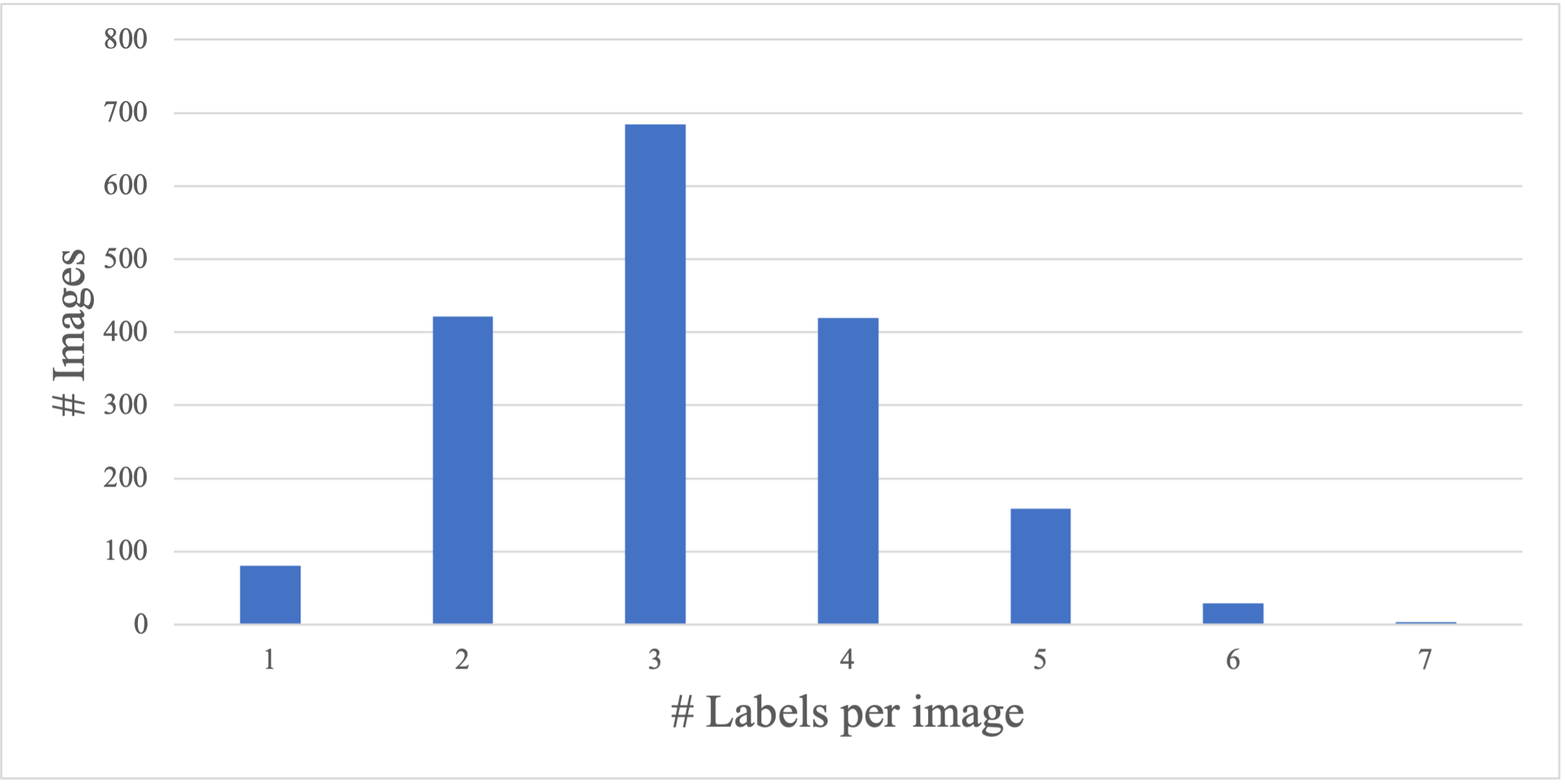}
\end{minipage}%
}%
\centering
\caption{Distributions of the proposed MAI dataset. (a) and (b) show the category-wised annotation distributions. (c) and (d) show the image distributions, with the number of labels in one image as the variable.}
\label{fig: data distribution}
\end{figure*}

Data is playing an especially critical role in enabling computers to interpret images as compositions of objects.
Based on the scenario described above, we created a new dataset named MAI, which contains images and ground-truth annotations for the single- to multi-label domain adaptation task.
The proposed MAI dataset contains two subsets, MAI-AID and MAI-UCM.
Besides, each subset is in pairs, which contains one single-label dataset and one multi-label dataset.
The statistical parameters of the dataset is detailed in Table. \ref{dataset:MAI} and Fig. \ref{fig: data distribution}.

\begin{table*}[!b]
\caption{Properties of The Proposed MAI Dataset} 
	\centering
	\begin{tabular}{|c|c|c|c|c|c|c|c|}  
		\hline
		& & & & & & & \\[-6pt]
		Dataset & Size & \# Images & \# Categories & \# Avg. Labels & \# Max. Labels & \# Min. Labels & Resolution \\ 
		\hline
		\hline
		& & & & \multicolumn{3}{c|}{} & \\[-6pt]
		MAI-AID-s & 1.7G & 7,050 & 20 &\multicolumn{3}{c|}{Single label annotation} & 600$\times$600\\ 
			\hline
		& & & & & & & \\[-6pt]
		MAI-AID-m & 198.0M& 3,239 & 20 & 3.73 & 9 & 1 & 512$\times$512 \\ 
		\hline
		& & & & \multicolumn{3}{c|}{} & \\[-6pt]
		MAI-UCM-s & 176.3M & 1,700 & 17 & \multicolumn{3}{c|}{Single label annotation} & 256$\times$256 \\
			\hline
		& & & & & & &\\[-6pt]
		MAI-UCM-m & 105.5M & 1,799 & 17 & 3.14 & 7 & 1 & 512$\times$512 \\ 
		\hline
	\end{tabular}
	\label{dataset:MAI}  
\end{table*}

\textbf{MAI-AID} As a large-scale aerial image dataset, AID is introduced in 2017\cite{xia2017aid}, which collects sample images from Google Earth imagery. 
The original AID dataset is made up of 30 categories, including 10,000 images.
7,050 images from 20 categories are collected from AID dataset and used as the single-label dataset named \textbf{MAI-AID-s}.
3,239 images with exactly the same 20 categories are collected from OpenStreetMap(OSM)\cite{OpenStreetMap} which is a collaborative project to create a free editable geographic database of the world and named \textbf{MAI-AID-m}.
The difference lies in the fact that the annotation of MAI-AID-s is single-label based and the annotation of MAI-AID-m is multi-label based.
All the images in MAI-AID-m are labelled by the specialists in the field of remote sensing image interpretation.
Some samples of MAI-AID-m dataset are exhibited in Fig. \ref{fig:AID examples}.

\textbf{MAI-UCM} UC Merced Land Use Dataset(UCM)\cite{yang2010bag} is an aerial image dataset whose images are manually extracted from the USGS National Map Urban Area Imagery collection for various urban areas around the country.
The original UCM dataset is made up of 21 categories, including 2,100 images.
1,700 images from 17 categories are collected from UCM dataset and used as the single-label dataset named \textbf{MAI-UCM-s}.
1,799 images with exactly the same 17 categories are collected from OSM and named \textbf{MAI-UCM-m}.
Similar to MAI-AID-m, the images of the MAI-UCM-m are also annotated with multi labels.
Fig. \ref{fig: UCM examples} exhibits four examples of MAI-UCM-m dataset with object level annotations.

Meanwhile, for multi-label datasets, there exist inner connections and correlations between different categories.
In Fig. \ref{fig:correlation}, the adjacent matrix of the two multi-label datasets are visualized in an intuitional way.
The line connecting two labels represents the connection and the width represents the correlation degree between two categories.
Among them, for instance, ``Residential`` and ``Parking Lot`` have wide correlations with other categories, while ``Runway`` only have a few and the closest correlation is with ``Airport``.

\begin{figure*}[ht]
  \centering
  \includegraphics[width=\linewidth]{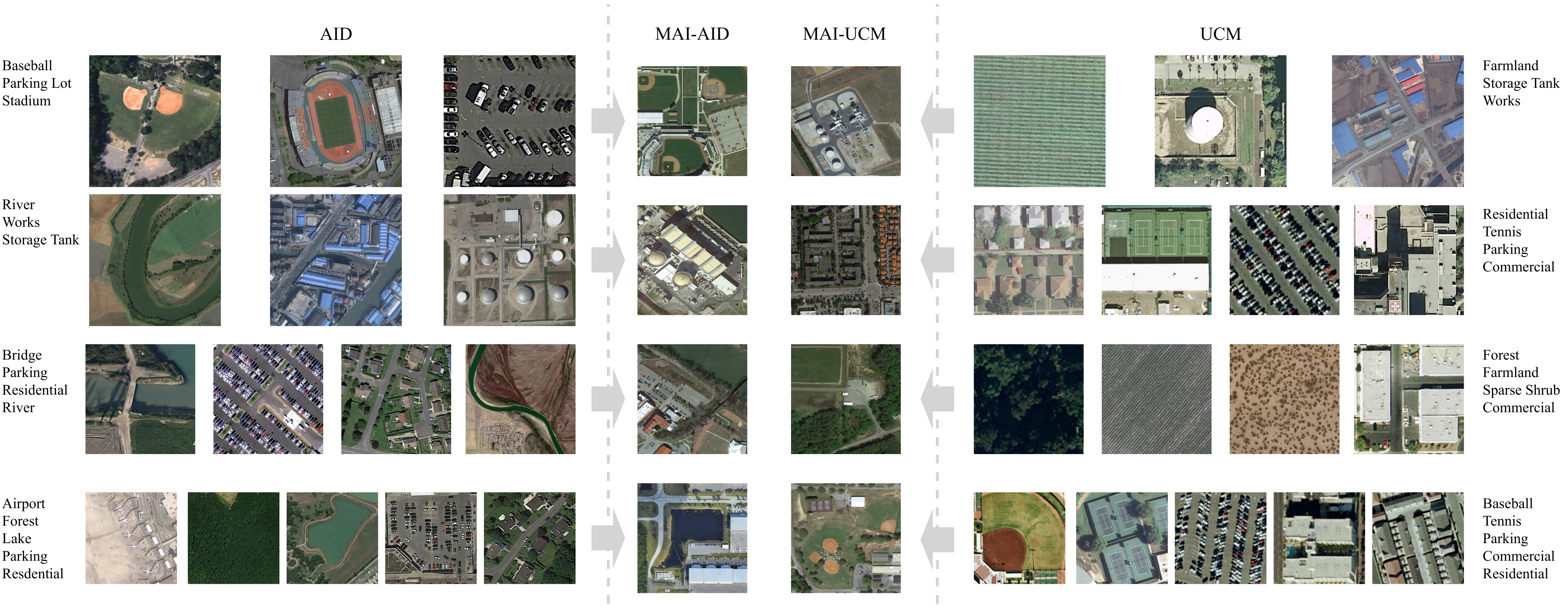}
  \caption{Image mappings of MAI-AID-s $\rightarrow $ MAI-AID-m and MAI-UCM-s $\rightarrow$ MAI-UCM-m in the proposed domain adaptation task.}
 \label{fig:example_data}
\end{figure*}

\subsubsection{Scene}
\emph{MAI-AID-s to MAI-AID-m adaptation:} For this task, we consider 20 classes for single- to multi-label domain adaptation, including
Airport, Baseball, Beach, Bridge, Commercial, Farmland, Forest, Lake, Parking Lot, Park, Port, Residential, River, Roundabout, Soccer Field, Stadium, Storage Tank, Train Station, Works, and Sea.
7,050 images from the MAI-AID-s dataset are used as the source domain, while 3,239 images from the MAI-AID-m dataset are used as the target domain. 
The number of source domain images in each class is imbalanced, ranging from 200 to 700. 
The input images are re-scaled to $512\times512$. Examples can be found in Fig. \ref{fig:example_data}.

\emph{MAI-UCM-s to MAI-UCM-m adaptation:} For this task, the datasets both have 17 classes, including Airport, Baseball, Beach, Commercial, Farmland, Forest, Golf Course, Parking Lot, Port, Residential, River, Runway, Sparse Shrub, Storage Tank, Tennis Court, Works, Sea. 
In total, 1,700 images from MAI-UCM-s are used as the source domain. 1,799 images from MAI-UCM-m are used as the target dataset. 
The input images are re-scaled to $224\times224$. Examples can be found in Fig. \ref{fig:example_data}.

\subsubsection{Setup}
A pre-trained ResNet-101\cite{resnet} is used as the backbone of the feature generator. 
For different datasets, the input images are randomly cropped and resized for data augmentation. SGD is used for network optimization. 
The momentum is set to be 0.9, with a decay of $10^{-4}$. The batch size is 4. 
The initial learning rate is 0.001 for the DWC branch, and 0.01 for the LWC branch. 
Both learning rates decay by a factor of 10 for every 30 epochs and 200 epochs.
The framework is implemented in PyTorch and trained with two 2080-TI GPUs.

\subsubsection{Evaluation Metrics}
In experiments, for performance evaluation, we report the average overall precision (OP), recall (OR), F1 (OF1), F2 (OF2); and the top-3 OP, OR, OF1, OF2 \cite{chen2019multi}.
Specifically, the F score (F1 when $\beta=1$ and F2 when $\beta=2$) is calculated using:

\begin{equation}
F_{\beta}=\left(1+\beta^{2}\right) \cdot \frac{\text { precision } \cdot \text { recall }}{\beta^{2} \cdot \text { precision }+\text { recall }}
\end{equation}

\subsection{Qualitative and Quantitative Comparisons}
In this section, the qualitative and quantitative results of the proposed task and method are detailed.
To fully explore the capacity of our proposed network, the order and the content of the experiments are designed on purpose.
The proposed task is single to multi-label domain adaptation, but there's no method reported in the literature for the proposed task yet.
So according to the structure of the proposed framework which is consist of two main modules, DWC branch and LWC branch, we have:

\begin{enumerate}
  \item comparison with the state-of-the-art methods for multi-label classification(e.g., the best performing methods KSSNet \cite{wang2019multi} and GCN-ASL \cite{benbaruch2020asymmetric} and the most representative method ML-GCN \cite{chen2019multi}) to prove the effectiveness of DWC branch.
  \item comparison with the state-of-the-art methods for domain adaptation approaches(e.g, the best performance unsupervised domain adaptation method HAFN/SAFN \cite{xu2019larger} and the most representative task-specific unsupervised domain adaptation(UDA) method MCD \cite{saito}) to prove the effectiveness of LWC branch.
  \item comparison with the state-of-the-art methods for partial label learning for multi-label classification(the PRODEN\cite{lv2020progressive} and the DNPL method \cite{seo2021power}).
\end{enumerate}


\begin{figure*}[ht]
  \centering
  \includegraphics[width=\linewidth]{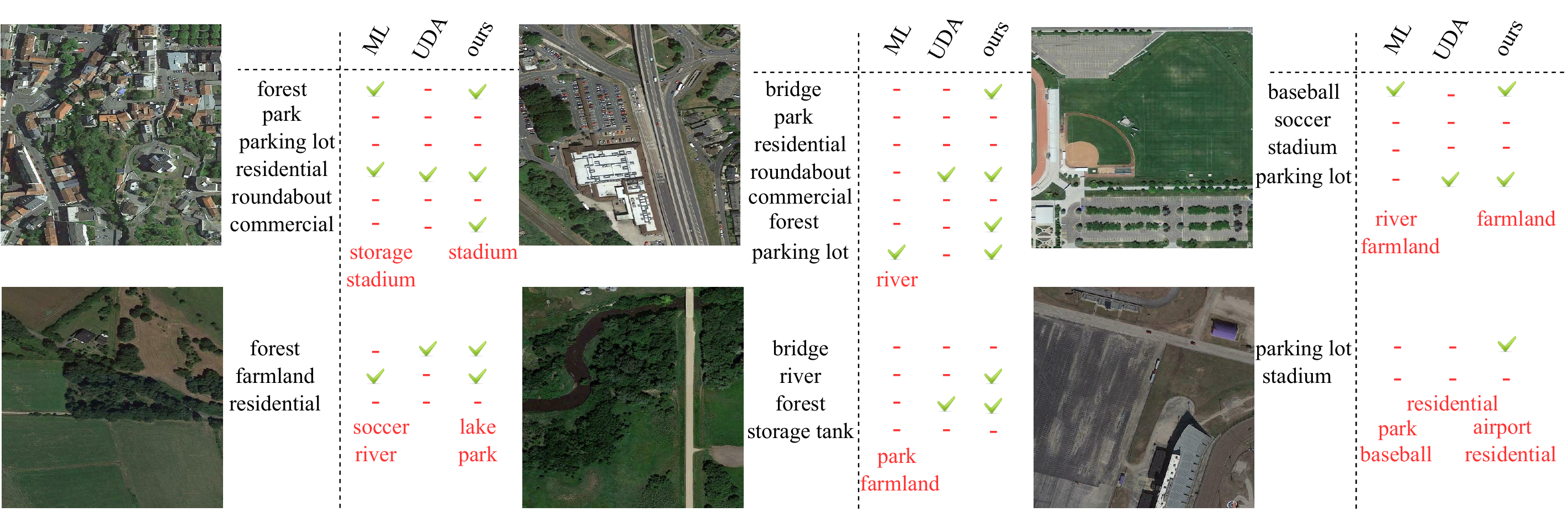}
  \caption{MAI-AID dataset: Visualization examples of the classification outputs. We use visualized results of ML-GCN \cite{chen2019multi} and MCD \cite{saito} for the results of multi-label classification (ML) and unsupervised domain adaptation (UDA) respectively.}
 \label{fig:example}
\end{figure*}

\begin{table*}[htbp]
\caption{MAI-AID Dataset: Classification Accuracy Comparisons with Different Multi-label Classification Methods}
\label{tb: AID to MAI-AID}
\centering
\begin{tabular}{c|llll|llll}
\hline
\multirow{2}{*}{} & \multicolumn{4}{c}{} & \multicolumn{4}{c}{} \\ [-6pt]  
\multirow{2}{*}{Method} & \multicolumn{4}{c}{All}                                                           & \multicolumn{4}{c}{Top 3}                                                         \\ \cline{2-9} 
 \multicolumn{1}{c|}{}& \multicolumn{1}{c|}{} & \multicolumn{1}{c|}{} &  \multicolumn{1}{c|}{}&  \multicolumn{1}{c|}{}&  \multicolumn{1}{c|}{}&  \multicolumn{1}{c|}{}& \multicolumn{1}{c|}{} \\ [-6pt]
                        & \multicolumn{1}{c|}{OP} & \multicolumn{1}{c|}{OR} & \multicolumn{1}{c|}{OF1} & \multicolumn{1}{c|}{OF2} & \multicolumn{1}{c|}{OP} & \multicolumn{1}{c|}{OR} & \multicolumn{1}{c|}{OF1} & \multicolumn{1}{c}{OF2} \\ \hline \hline
\multicolumn{1}{c|}{} & \multicolumn{1}{c}{} & \multicolumn{1}{c}{} & \multicolumn{1}{c}{} & \multicolumn{1}{c|}{} & \multicolumn{1}{c}{} & \multicolumn{1}{c}{} & \multicolumn{1}{c}{} & \\ [-7pt]
                KSSNet \cite{wang2019multi} & 0.2907 & 0.1616  & 0.2077 & 0.1774 & 0.3336 & 0.1303 & 0.1874 & 0.1484 \\
\multicolumn{1}{c|}{} & \multicolumn{1}{c}{} & \multicolumn{1}{c}{} & \multicolumn{1}{c}{} & \multicolumn{1}{c|}{} & \multicolumn{1}{c}{} & \multicolumn{1}{c}{} & \multicolumn{1}{c}{} & \\ [-7pt]
                GCN-ASL \cite{benbaruch2020asymmetric} & 0.2026  & 0.1615 & 0.1797 & 0.1683 & 0.2742  & 0.0389  & 0.0681 & 0.0470 \\
\multicolumn{1}{c|}{} & \multicolumn{1}{c}{} & \multicolumn{1}{c}{} & \multicolumn{1}{c}{} & \multicolumn{1}{c|}{} & \multicolumn{1}{c}{} & \multicolumn{1}{c}{} & \multicolumn{1}{c}{} & \\ [-7pt]
                ML-GCN \cite{chen2019multi}  & 0.3831 & 0.0452 & 0.0809 & 0.0549 & 0.3837 & 0.0447 & 0.0801 & 0.0543 \\ \hline \hline
\multicolumn{1}{c|}{} & \multicolumn{1}{c}{} & \multicolumn{1}{c}{} & \multicolumn{1}{c}{} & \multicolumn{1}{c|}{} & \multicolumn{1}{c}{} & \multicolumn{1}{c}{} & \multicolumn{1}{c}{} & \\ [-7pt]
                SCIDA    & \textbf{0.5432}    & 0.2230    & 0.3162    & 0.2528    & \textbf{0.5496}    & 0.2196    & 0.3138    & 0.2496 \\
\multicolumn{1}{c|}{} & \multicolumn{1}{c}{} & \multicolumn{1}{c}{} & \multicolumn{1}{c}{} & \multicolumn{1}{c|}{} & \multicolumn{1}{c}{} & \multicolumn{1}{c}{} & \multicolumn{1}{c}{} & \\ [-7pt]
        SCIDA(opt-$\delta$)  & 0.4474    & \textbf{0.3242}    & \textbf{0.3760}    & \textbf{0.3431}    & 0.4725    & \textbf{0.3185}    & \textbf{0.3805}    & \textbf{0.3407} \\ \hline
\end{tabular}
\end{table*}

\begin{table*}[ht]
\caption{MAI-UCM Dataset: Classification Accuracy Comparisons with Different Multi-label Classification Methods}
\label{tb: UCM to MAI-UCM}
\centering
\begin{tabular}{c|llll|llll}
\hline
\multirow{2}{*}{} & \multicolumn{4}{c}{}  & \multicolumn{4}{c}{} \\ [-6pt]  
\multirow{2}{*}{Method} & \multicolumn{4}{c}{All} & \multicolumn{4}{c}{Top 3} \\ \cline{2-9} 
 \multicolumn{1}{c|}{}& \multicolumn{1}{c|}{} & \multicolumn{1}{c|}{} &  \multicolumn{1}{c|}{}&  \multicolumn{1}{c|}{}&  \multicolumn{1}{c|}{}&  \multicolumn{1}{c|}{}& \multicolumn{1}{c|}{} \\ [-6pt]
                        & \multicolumn{1}{c|}{OP} & \multicolumn{1}{c|}{OR} & \multicolumn{1}{c|}{OF1} & \multicolumn{1}{c|}{OF2} & \multicolumn{1}{c|}{OP} & \multicolumn{1}{c|}{OR} & \multicolumn{1}{c|}{OF1} & \multicolumn{1}{c}{OF2} \\ \hline \hline
\multicolumn{1}{c|}{} & \multicolumn{1}{c}{} & \multicolumn{1}{c}{} & \multicolumn{1}{c}{} & \multicolumn{1}{c|}{} & \multicolumn{1}{c}{} & \multicolumn{1}{c}{} & \multicolumn{1}{c}{} & \\ [-7pt]
                KSSNet \cite{wang2019multi} & 0.2817  & 0.1804  & 0.2199 & 0.1944 & 0.2829 & 0.1769 & 0.2177 & 0.1912    \\
\multicolumn{1}{c|}{} & \multicolumn{1}{c}{} & \multicolumn{1}{c}{} & \multicolumn{1}{c}{} & \multicolumn{1}{c|}{} & \multicolumn{1}{c}{} & \multicolumn{1}{c}{} & \multicolumn{1}{c}{} & \\ [-7pt]
                GCN-ASL \cite{benbaruch2020asymmetric} & 0.1844 & 0.3200 & 0.2340 & 0.2790 & 0.1579 & 0.0487 & 0.0745 & 0.0565 \\
\multicolumn{1}{c|}{} & \multicolumn{1}{c}{} & \multicolumn{1}{c}{} & \multicolumn{1}{c}{} & \multicolumn{1}{c|}{} & \multicolumn{1}{c}{} & \multicolumn{1}{c}{} & \multicolumn{1}{c}{} & \\ [-7pt]
                MC-GCN \cite{chen2019multi} & \textbf{0.3585} & 0.0404 & 0.0726 & 0.0491 & \textbf{0.4127} & 0.0356 & 0.0656 & 0.0436 \\ \hline \hline
\multicolumn{1}{c|}{} & \multicolumn{1}{c}{} & \multicolumn{1}{c}{} & \multicolumn{1}{c}{} & \multicolumn{1}{c|}{} & \multicolumn{1}{c}{} & \multicolumn{1}{c}{} & \multicolumn{1}{c}{} & \\ [-7pt]
                SCIDA    & 0.3358 & 0.3105 & 0.3227 & 0.3153 & 0.3412 & 0.2995 & 0.3190 & 0.3070 \\
\multicolumn{1}{c|}{} & \multicolumn{1}{c}{} & \multicolumn{1}{c}{} & \multicolumn{1}{c}{} & \multicolumn{1}{c|}{} & \multicolumn{1}{c}{} & \multicolumn{1}{c}{} & \multicolumn{1}{c}{} & \\ [-7pt]
        SCIDA(opt-$\delta$)  & 0.3371 & \textbf{0.3219} & \textbf{0.3293} & \textbf{0.3248} & 0.3380 & \textbf{0.3192} & \textbf{0.3284} & \textbf{0.3228}\\ \hline
\end{tabular}
\end{table*}

An intuitive comparison can be found in Fig. \ref{fig:example}.
In this figure, 6 images with different number of labels ranging from 2 to 7 are chosen for demonstration.
The multi label method could have multiple prediction. 
But because of the existence of the domain gap between the source and the target dataset, the results are not satisfying.
On the other hand, UDA method shows a higher precision on the prediction than ML, but because of the its own limitation, it cannot learn the correlation between the categories of the target dataset.
The proposed method shows a much higher accuracy on predicting the multiple labels in the proposed task.

\noindent\textbf{Comparisons with multi-label classification methods.}\quad 
Quantitative results are reported in Table. \ref{tb: AID to MAI-AID} and \ref{tb: UCM to MAI-UCM}. 
The results of SCIDA and SCIDA with optimized-$\delta$ are reported separately. 
All comparison methods are trained on the annotated single label data (MAI-AID-s/MAI-UCM-s) and tested directly on the multi-label data (MAI-AID-m/MAI-UCM-m).

For the MAI-AID-s to MAI-AID-m task, it is evident that the proposed SCIDA method provides superior classification performances under all metrics. 
For SCIDA with optimized-$\delta$, the OP drops a bit because more related correlation estimation is made, while OR/OF1/OF2 improve significantly. 
For the MAI-UCM-s to MAI-UCM-m task, our proposed method outperforms other comparison methods under all metrics except OP. 
With a similar OP accuracy, the OR/OF1/OF2 are improved by about 30\%. 
For both tasks, the optimized SCIDA consistently achieves better performances than the regular SCIDA. 
Based on the performance comparisons with multi-label classification methods (without domain adaptation) in Table. \ref{tb: AID to MAI-AID} and \ref{tb: UCM to MAI-UCM}, we can verify that the correlation between single- and multi-label data learned by domain adaptation is significant.

\begin{table*}[ht]
\caption{MAI-AID Dataset: Classification Accuracy Comparisons with Different Domain Adaptation Methods}
\label{tb: AID to MAI-AID adaptation}
\centering
\begin{tabular}{c|llll|llll}
\hline
\multirow{2}{*}{} & \multicolumn{4}{c}{}  & \multicolumn{4}{c}{} \\ [-6pt]  
\multirow{2}{*}{Method} & \multicolumn{4}{c}{All} & \multicolumn{4}{c}{Top 3} \\ \cline{2-9} 
 \multicolumn{1}{c|}{}& \multicolumn{1}{c|}{} & \multicolumn{1}{c|}{} &  \multicolumn{1}{c|}{}&  \multicolumn{1}{c|}{}&  \multicolumn{1}{c|}{}&  \multicolumn{1}{c|}{}& \multicolumn{1}{c|}{} \\ [-6pt]
                        & \multicolumn{1}{c|}{OP} & \multicolumn{1}{c|}{OR} & \multicolumn{1}{c|}{OF1} & \multicolumn{1}{c|}{OF2} & \multicolumn{1}{c|}{OP} & \multicolumn{1}{c|}{OR} & \multicolumn{1}{c|}{OF1} & \multicolumn{1}{c}{OF2} \\ \hline \hline
\multicolumn{1}{c|}{} & \multicolumn{1}{c}{} & \multicolumn{1}{c}{} & \multicolumn{1}{c}{} & \multicolumn{1}{c|}{} & \multicolumn{1}{c}{} & \multicolumn{1}{c}{} & \multicolumn{1}{c}{} & \\ [-7pt]
SAFN  \cite{xu2019larger}  & 0.4542 & 0.1216 & 0.1918 & 0.1425 & 0.4542 & 0.1216 & 0.1918 & 0.1425 \\
\multicolumn{1}{c|}{} & \multicolumn{1}{c}{} & \multicolumn{1}{c}{} & \multicolumn{1}{c}{} & \multicolumn{1}{c|}{} & \multicolumn{1}{c}{} & \multicolumn{1}{c}{} & \multicolumn{1}{c}{} & \\ [-7pt]
HAFN  \cite{xu2019larger}  & 0.4281 & 0,1147 & 0.1809 & 0.1344 & 0.4281 & 0.1147 & 0.1809 & 0.1344 \\
\multicolumn{1}{c|}{} & \multicolumn{1}{c}{} & \multicolumn{1}{c}{} & \multicolumn{1}{c}{} & \multicolumn{1}{c|}{} & \multicolumn{1}{c}{} & \multicolumn{1}{c}{} & \multicolumn{1}{c}{} & \\ [-7pt]
MCD \cite{saito}    & 0.3327 & 0.0891 & 0.1406 & 0.1044 & 0.3327 & 0.0891 & 0.1406 & 0.1044 \\ \hline \hline
\multicolumn{1}{c|}{} & \multicolumn{1}{c}{} & \multicolumn{1}{c}{} & \multicolumn{1}{c}{} & \multicolumn{1}{c|}{} & \multicolumn{1}{c}{} & \multicolumn{1}{c}{} & \multicolumn{1}{c}{} & \\ [-7pt]
SCIDA    & \textbf{0.5432}    & 0.2230    & 0.3162    & 0.2528 & \textbf{0.5496}    & 0.2196    & 0.3138    & 0.2496   \\ 
\multicolumn{1}{c|}{} & \multicolumn{1}{c}{} & \multicolumn{1}{c}{} & \multicolumn{1}{c}{} & \multicolumn{1}{c|}{} & \multicolumn{1}{c}{} & \multicolumn{1}{c}{} & \multicolumn{1}{c}{} & \\ [-7pt]
SCIDA(opt-$\delta$)  & 0.4474    & \textbf{0.3242}    & \textbf{0.3760}    & \textbf{0.3431}  & 0.4725    & \textbf{0.3185}    & \textbf{0.3805}    & \textbf{0.3407}   \\ \hline \hline
\end{tabular}
\end{table*}

\begin{table*}[htbp]
\renewcommand{\arraystretch}{1.3}
\caption{MAI-AID Dataset: Classification Accuracy Comparisons with Partial-label Learning Methods}
\label{tb: partial_1}
\centering
\begin{tabular}{c|llll|llll}
\hline
\multirow{2}{*}{Method} & \multicolumn{4}{c}{All}                                                           & \multicolumn{4}{c}{Top 3}                                                         \\ \cline{2-9} 
                        & \multicolumn{1}{c|}{OP} & \multicolumn{1}{c|}{OR} & \multicolumn{1}{c|}{OF1} & \multicolumn{1}{c|}{OF2} & \multicolumn{1}{c|}{OP} & \multicolumn{1}{c|}{OR} & \multicolumn{1}{c|}{OF1} & \multicolumn{1}{c}{OF2} \\ \hline \hline
                PRODEN \cite{lv2020progressive} & 0.1559  & 0.0422 & 0.0664 & 0.0494 & 0.1559  & 0.0422 & 0.0664 & 0.049 \\
                DNPL    \cite{seo2021power}  & 0.1031 & 0.0402 & 0.0578 & 0.0458 & 0.1031 & 0.0402 & 0.0578 & 0.0458  \\ \hline \hline
                SCIDA    & \textbf{0.5432}    & 0.2230    & 0.3162    & 0.2528    & \textbf{0.5496}    & 0.2196    & 0.3138    & 0.2496 \\
        SCIDA(opt-$\delta$)  & 0.4474    & \textbf{0.3242}    & \textbf{0.3760}    & \textbf{0.3431}    & 0.4725    & \textbf{0.3185}    & \textbf{0.3805}    & \textbf{0.3407} \\ \hline
\end{tabular}
\end{table*}

\noindent\textbf{Comparisons with domain adaptation methods.}\quad 
Since the resolution of the MAI-UCM images is quite low, i.e., only 1/4 of that of MAI-AID images, almost all existing domain adaptation methods fail on the MAI-UCM dataset, except our proposed method with self-correction. 
Therefore, we only report the performance comparisons for the MAI-AID-s to MAI-AID-m task. 
As shown in Table. \ref{tb: AID to MAI-AID adaptation}, the proposed methods significantly outperform comparison methods under all performance metrics. 
Especially the super performances of the optimized SCIDA (with opt-$\delta$) clearly demonstrate the effectiveness of our proposed self-correction module (LWC branch).

\noindent\textbf{Comparison with partial label learning methods.}\quad
Partial-label learning(PLL) is a typical weakly supervised learning problem, where each training instance consists of a data and a set of candidate labels containing a unique ground truth label. 
The task setting is the same as our source domain dataset. 
The difference is that our target domain consists of an uncertain number of labels.
The candidate labels are set to be the number of the classes in the experiments.
Besides, all the algorithms have the same training, testing dataset and the same number of annotations.
Table. \ref{tb: partial_1} and \ref{tb: partial_2} exhibit the results on MAI-AID and MAI-UCM dataset respectively.
We can observe that our model surpasses the two competitors on both datasets.
Specifically, compared with the two partial label learning methods, the proposed method improves the F1 and F2 score by more than 0.3. 


\begin{figure*}[htbp]
\centering
\subfigure[Training curve when $\delta=0.15$.]{
\begin{minipage}[t]{0.5\linewidth}
\centering
\includegraphics[width=0.9\textwidth]{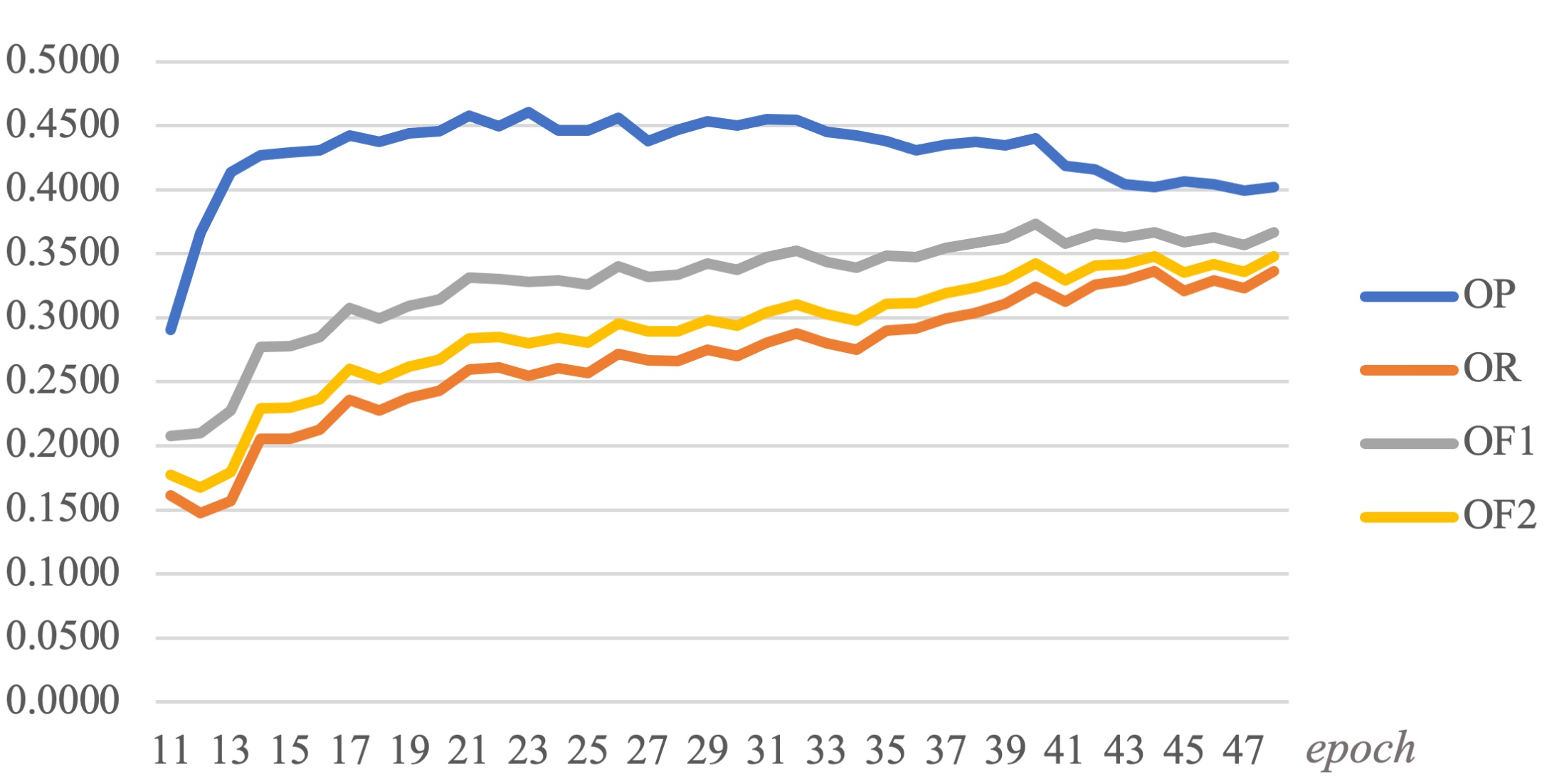}
\end{minipage}%
}%
\subfigure[Training curve when $\delta=0.2$.]{
\begin{minipage}[t]{0.5\linewidth}
\centering
\includegraphics[width=0.9\textwidth]{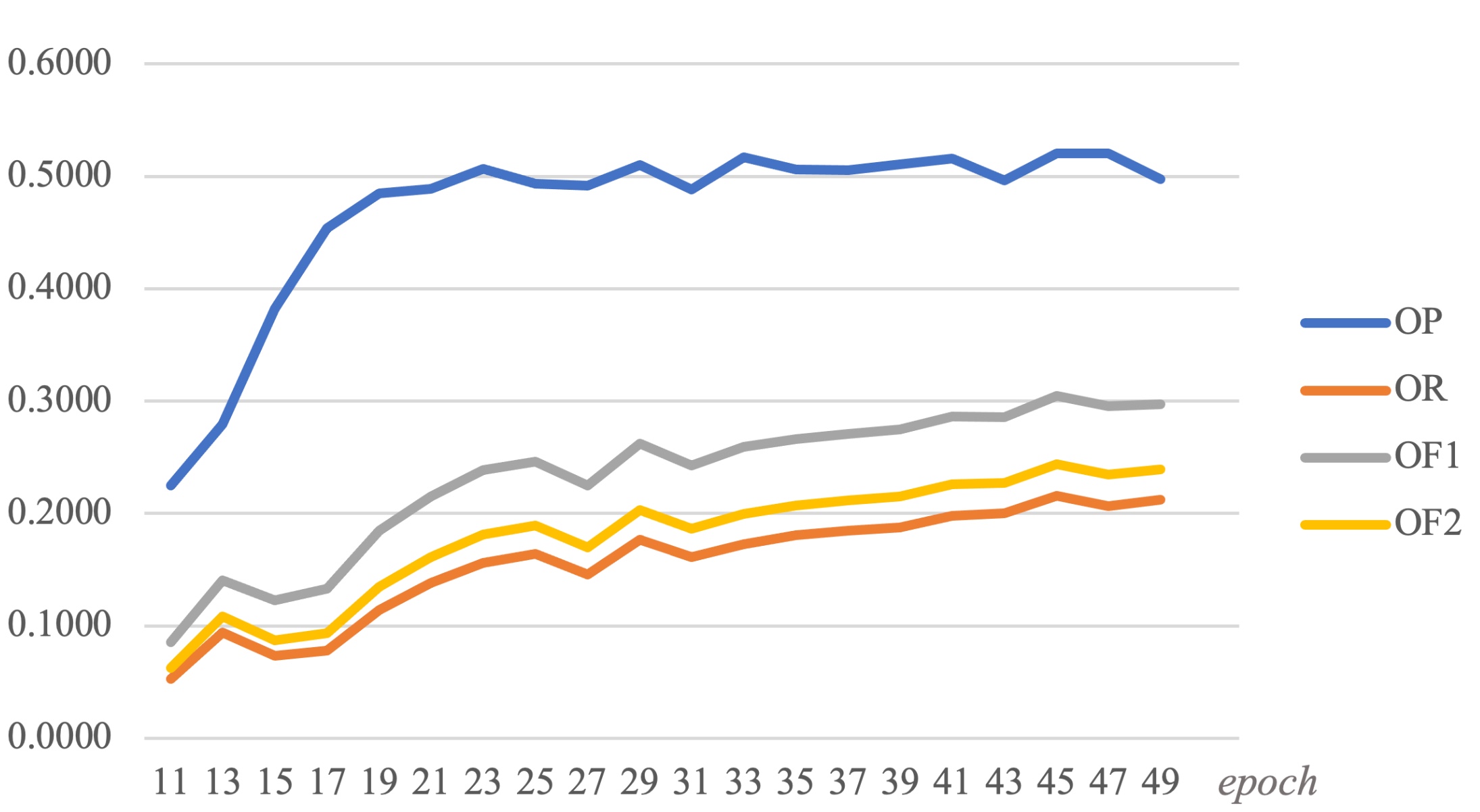}
\end{minipage}%
}%
\centering
\caption{Ablation study: Training curves of SCIDA with different values of $\delta$. The evaluation measures OP, OR, OF1 and OF2 of every epoch are logged until the model converges and is stable. To better show the figures, the initial 10 epochs are skipped when plotting.}
\label{fig: ablation4}
\end{figure*}

\begin{table*}[htbp]
\renewcommand{\arraystretch}{1.3}
\caption{MAI-UCM Dataset: Classification Accuracy Comparisons with Partial-label Learning Methods}
\label{tb: partial_2}
\centering
\begin{tabular}{c|llll|llll}
\hline
\multirow{2}{*}{Method} & \multicolumn{4}{c}{All} & \multicolumn{4}{c}{Top 3} \\ \cline{2-9} 
                        & \multicolumn{1}{c|}{OP} & \multicolumn{1}{c|}{OR} & \multicolumn{1}{c|}{OF1} & \multicolumn{1}{c|}{OF2} & \multicolumn{1}{c|}{OP} & \multicolumn{1}{c|}{OR} & \multicolumn{1}{c|}{OF1} & \multicolumn{1}{c}{OF2} \\ \hline \hline
                PRODEN \cite{lv2020progressive} & 0.1421  & 0.0560 & 0.0803 & 0.0637 & 0.1421  & 0.0560 & 0.0803 & 0.0637 \\
                DNPL    \cite{seo2021power}  & 0.1143 & 0.0478 & 0.0674 & 0.0541 & 0.1143 & 0.0478 & 0.0674 & 0.0541  \\ \hline \hline
                SCIDA    & \textbf{0.3358} & 0.3105 & 0.3227 & 0.3153 & \textbf{0.3412} & 0.2995 & 0.3190 & 0.3070 \\
        SCIDA(opt-$\delta$)  & 0.3371 & \textbf{0.3219} & \textbf{0.3293} & \textbf{0.3248} & 0.3380 & \textbf{0.3192} & \textbf{0.3284} & \textbf{0.3228}\\ \hline
\end{tabular}
\end{table*}

\subsection{Ablation Studies}
In this section, we perform ablation studies on the other parts of the framework which are not covered in the experiments above.
Based on the task of MAI-AID-s to MAI-AID-m adaptation, the ablation studies are separated into the following aspects:

\begin{enumerate}
	\item the effectiveness of GCN module
	\item the effectiveness of LWC branch
	\item the comparison of different loss functions
	\item the influence of the parameter $\delta$
	\item stability of the proposed framework
	\item the results under the same annotation budget
\end{enumerate}

\noindent\textbf{GCN module.}\quad
In this study, we compare the results of the proposed framework SCIDA with and without GCN module to verify the importance of GCN.
The results are shown in Table. \ref{tb: ablation1}.
According to the result, we could verify that, by introducing GCN module, the performance of SCIDA is much improved.

\begin{table*}[ht]
\renewcommand{\arraystretch}{1.3}
\caption{Ablation Study: Classification Accuracy Comparisons of the Proposed SCIDA with and without GCN.}
\label{tb: ablation1}
\centering
\begin{tabular}{c|llll|llll}
\hline
\multirow{2}{*}{Method} & \multicolumn{4}{c}{All} & \multicolumn{4}{c}{Top 3} \\ \cline{2-9} 
                        & \multicolumn{1}{c|}{OP} & \multicolumn{1}{c|}{OR} & \multicolumn{1}{c|}{OF1} & \multicolumn{1}{c|}{OF2} & \multicolumn{1}{c|}{OP} & \multicolumn{1}{c|}{OR} & \multicolumn{1}{c|}{OF1} & \multicolumn{1}{c}{OF2} \\ \hline \hline
                SCIDA(None-GCN)    & 0.3423 & \textbf{0.3688} & 0.3551 & \textbf{0.3641} & 0.3652 & 0.2945 & 0.3260 & 0.3060 \\
        		SCIDA 							& \textbf{0.4474}    & 0.3242    & \textbf{0.3760}   & 0.3431 & \textbf{0.4725}    & \textbf{0.3185}    & \textbf{0.3805}    & \textbf{0.3407}\\ \hline
\end{tabular}
\end{table*}

\noindent\textbf{With/without LWC branch.}\quad
In this part, we generally evaluate the effectiveness of the proposed LWC branch. 
By deleting the LWC branch directly and only optimizing the DWC branch, we can get around 0.14 for OF1 and 0.10 for OF2. 
In comparison, SCIDA gets 0.38 for OF1 and 0.34 for OF2. This result verifies that the proposed LWC module is quite necessary for our framework.

\begin{table}[!b]
\renewcommand{\arraystretch}{1.3}
\caption{Ablation Study: Comparisons When Using Different Loss Functions}
\label{tb: ablation-lossfunction}
\centering
\begin{tabular}{c|llll}
\hline
\multicolumn{1}{c|}{\multirow{2}{*}{Method}} & \multicolumn{4}{c}{All}   \\ \cline{2-5} 
\multicolumn{1}{c|}{}                        & \multicolumn{1}{c|}{OP}      & \multicolumn{1}{c|}{OR}     & \multicolumn{1}{c|}{OF1}   & \multicolumn{1}{c}{OF2}    \\ \hline \hline
BCE Loss                                          &  0.1747     &  0.2475    &  0.2048    &   0.2285   \\ \hline
wFL                                           &   \textbf{0.4546}    &   \textbf{0.2878}   &   \textbf{0.3524}   &   \textbf{0.3106}   \\ \hline \hline
                                              & \multicolumn{4}{c}{Top-3} \\ \hline
BCE Loss                                          &   0.1480   &   0.0764   &   0.1008   &   0.0846   \\ \hline
wFL                                           &   \textbf{0.4654}    &   \textbf{0.2806}   &  \textbf{0.3501}    &   \textbf{0.3048}   \\ \hline
\end{tabular}
\end{table}

\noindent\textbf{BCE loss vs weight focal loss (wFL).}\quad
In this part, we evaluate different loss functions for step-1 training. 
Specifically, we investigate two loss functions, including the widely used BCE loss and the proposed wFL.
To make a fair comparison, hyper-parameters under these two loss functions are tuned specifically to achieve the best performance.
The results of the two loss functions are both selected when the model is converged and stable.
Table. \ref{tb: ablation-lossfunction} shows the results using different loss functions on the MAI-AID-s to MAI-AID-m task. 
We can see that the wFL clearly yields better accuracy under all performance metrics.

\begin{figure}[!htb]
  \centering
  \includegraphics[width=\linewidth]{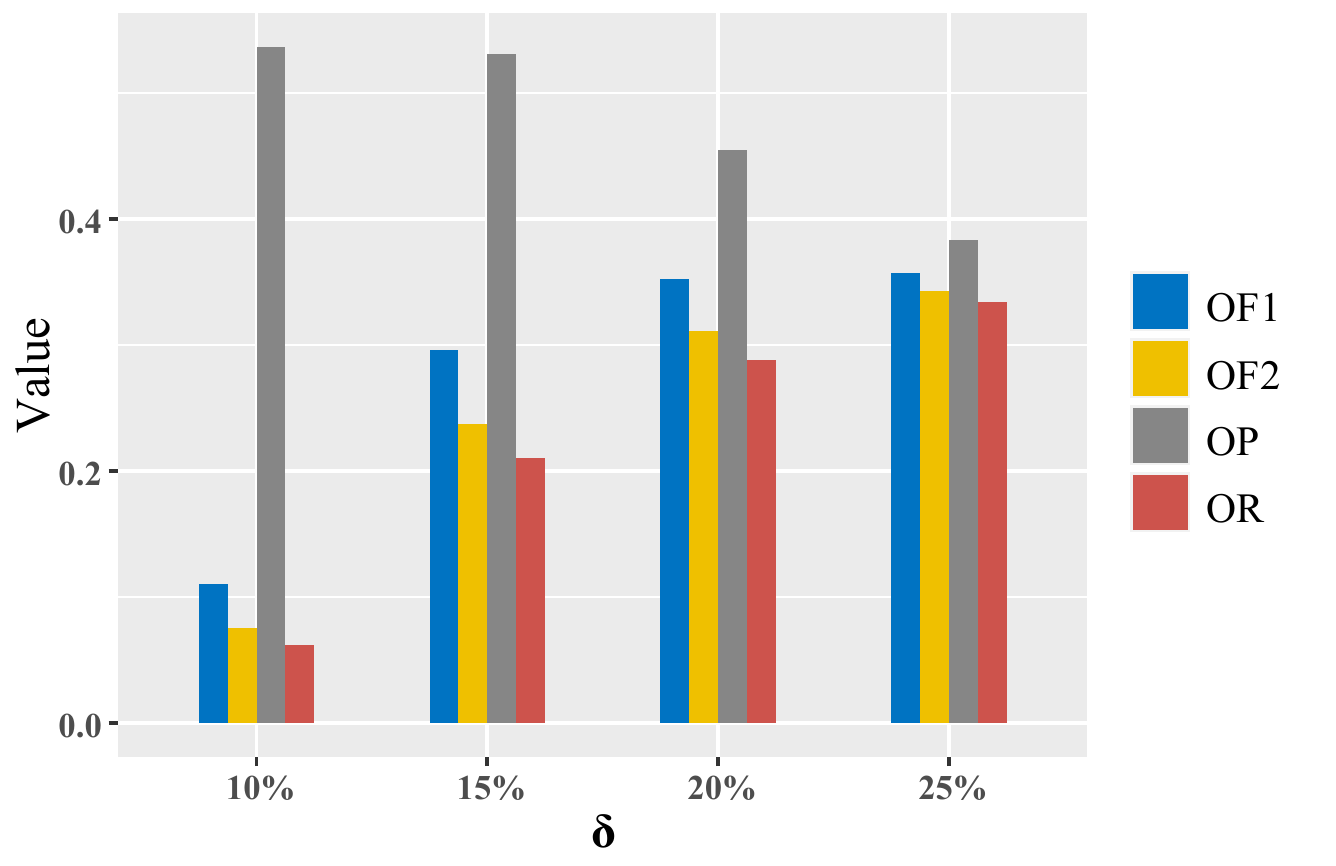}
  \caption{Ablation study: Performance evaluation when using different $\delta$.}
  \label{fig:ablation_chart1}
\end{figure}

\noindent\textbf{Different values of $\delta$ for LWC branch.}\quad
To explore the effects of $\delta$ on classification performance, we consider different values of $\delta$, ranging from 10\% to 25\%, as depicted in Fig. \ref{fig:ablation_chart1}. 
We can observe that, when $\delta$ is set as 20\%, the performance is the best. 
If $\delta$ is too small, the correlation learning in the GNN will be slow and the training of the branch will be insufficient; while when $\delta$ is set larger than $20\%$, redundant and useless connections will seriously affect the LWC branch, resulting in a worse overall performance. 
Therefore, we empirically set $\delta$ as 20\%.

\noindent\textbf{Model stability.} \quad
The proposed framework SCIDA is consist of several components.
The stability of the framework is quite critical.
In fact, the two-stage training manner introduced in Sec. \ref{sec:ModelTraining} could ensure the stability of the model to a great extent.
To verify that, the training curve of SCIDA under different $\delta$ values  are plotted, as shown in Fig. \ref{fig: ablation4}.
The proposed method is quite stable under different conditions.

\noindent\textbf{Same budget of annotations.} \quad
In this experiment, we want to demonstrate that with the same number of annotations, when compared with the method directly trained on the multi-label dataset, the proposed transfer method can yield comparable results. 
First, the proposed SCIDA uses 5,000 images from the AID dataset for training. 
In comparison, we train the state-of-the-art multi-label classification method ML-GCN using the target MAI-AID images directly. 
As the average number of annotations per image for MAI-AID is 3.4, 1,500 images (with $3.4\times 1500 \approx 5000$ labels) are chosen randomly for training and the remaining are used for testing. 
With the same number of annotations as described above, compared with the 0.4355 (OF1) and 0.3820 (OF2) of ML-GCN which is trained directly on the target data, SCIDA could achieve comparable results as 0.3673(OF1) and 0.3493 (OF2). 
It is worth emphasizing that the proposed SCIDA only uses publicly available annotated single-label images. 
The results suggest that for model training, prior knowledge from publicly available single-label images can be an efficient alternative to manual annotations of multi-label images.

\section{Conclusion}
In this paper, we propose a novel framework for single-label to multi-label aerial scene transfer. 
The proposed SCIDA model integrates self-correction to domain adaptation. This model can be applied to large scale, unlabeled and unconstrained aerial images. 
The model is trained in a two-stage manner. Our reported multi-label classification results in the target domain demonstrate the effectiveness of the proposed model. 
A new multi-label aerial image (MAI) dataset is collected and used for experiments. 
For future work, we will extend the proposed SCIDA model to more challenging image types and applications.

{
\bibliographystyle{IEEEtran}
\bibliography{egbib}
}
\end{document}